%% file: main.tex
\pdfoutput=1
\documentclass{article}

\PassOptionsToPackage{numbers,compress}{natbib}
\usepackage[main,final]{neurips/neurips_2026}

\usepackage[utf8]{inputenc} %
\usepackage[T1]{fontenc}    %
\usepackage{hyperref}       %
\usepackage{url}            %
\usepackage{booktabs}       %
\usepackage{amsfonts}       %
\usepackage{nicefrac}       %
\usepackage{microtype}      %
\usepackage{xcolor}         %

\usepackage{algorithm}
\usepackage{kotex}
\usepackage{multirow}
\usepackage{makecell}
\usepackage{graphicx}
\usepackage{adjustbox}
\usepackage{float}
\usepackage{wrapfig}
\usepackage{colortbl}
\usepackage[endLComment=,italicComments=false]{algpseudocodex}
\usepackage{enumitem}

\usepackage{amsmath}
\usepackage{amssymb}
\usepackage{mathtools}
\usepackage{amsthm}

\usepackage{pifont}%
\usepackage{xspace}
\usepackage{dsfont}

\usepackage{amsmath,amsfonts}
\definecolor{myred}{HTML}{F54254}
\definecolor{myorange}{HTML}{FFB135}
\definecolor{mygreen}{HTML}{163611}
\definecolor{myblue}{HTML}{598BE7}
\definecolor{mypurple}{HTML}{9A1C6B}
\definecolor{plgray}{HTML}{999999}

\newcommand{\ours}{\texttt{RecPo}\xspace}
\usepackage{lipsum}
\usepackage{comment}

\title{Reconfiguring Procedural Knowledge for Compositional Robot Skill Adaptation}
\author{%
  Daehee Lee$^{\spadesuit}$ \quad
  Dongsu Lee$^{\diamondsuit}$ \quad
  Sanghyun Ahn$^{\spadesuit}$\thanks{Sanghyun Ahn is currently a visiting student at Carnegie Mellon University.} \quad
  Minjong Yoo$^{\clubsuit}$ \quad
  Honguk Woo$^{\spadesuit}$\thanks{Corresponding author: Honguk Woo (hwoo@skku.edu).} \\
  $^{\spadesuit}$Sungkyunkwan University \quad
  $^{\diamondsuit}$University of Texas at Austin \quad
  $^{\clubsuit}$Samsung Electronics \\
  \texttt{dulgi7245@skku.edu, hwoo@skku.edu}
}

\begin{document}

\maketitle

\begin{abstract} 
General-purpose behavior models increasingly internalize broad procedural knowledge during training, making their adaptation to new goals and skill interfaces less a problem of learning entirely new behaviors than of reconfiguring existing competence.
We study this capability as \textit{\textbf{reconfigurability}}, the ability to realize new behaviors by cost-efficiently restructuring existing components and relations under \textit{\textbf{sparse governing information}}, where task specifications provide high-level goals and procedure specifications expose a limited set of reusable factors, leaving the concrete bindings and compositions required for execution underdetermined.
We propose \ours, a framework that represents skills task-agnostically through multiple procedure views and materializes executable skill sequences by composing stored transitions to realize task-specific changes.
RecPo maintains a procedure knowledge base of skill-induced transitions, with each view exposing a reusable factor and preserving local skill connectivity. At deployment, it interprets task specifications through the same views and resolves the resulting compositional queries over the stored relations to bind skills for execution.
On long-horizon tasks in an extended Franka Kitchen benchmark, \ours{} achieves $91.9\%$ ordered completion with $1.0$~ms of task inference per episode across varying procedure support and task-record sparsity. Real-robot studies further demonstrate adaptation through local memory edits without task-specific retraining.
\end{abstract}

\section{Introduction}
Reconfigurability is a core systems-level capability for lifelong embodied agents deployed in open-ended physical environments, enabling cost-effective adaptation under the constrained computational and physical resources available on-site~\cite{defCRL2023, kwon2023lifelearner, syed2025expres}.
Since design-time knowledge cannot cover every case~\cite{madani1999undecidability} and runtime exposes only a partial and shifting subset of deployment conditions, agents must locally update existing competence as user specifications over behaviors and tasks change. These limitations compound for embodied agents, whose deployments take place in physical environments where spatial layouts, object properties, and physical dynamics vary across sites and shift over time.

Modern Vision-Language-Action (VLA) models~\cite{BlackK-RSS-25,pmlr-v305-black25a,nvidia2025gr00tn1,physicalintelligence2025pistar06,physicalintelligence2026pi07} consolidate broad procedural knowledge at training time, leaving deployment to local adaptation. Performing such adaptation in practice, however, remains difficult: full retraining is costly and prone to catastrophic forgetting~\cite{hancock2025actions, mai2026survey}, while ad hoc conditioning offers only transient adjustments~\cite{liu2026vls}.
Therefore, reconfiguration is not only about adaptation but also about externalizing this procedural knowledge into a form governable through its environment-grounded effects, to enable cost-effective updates at deployment
~\cite{liang2022search, schmied2025retrievalaugmented, qin2026harnessing, xia2026grasp, zhou2026externalization}.

Recent efforts toward reconfigurability increasingly rely on external memories of prior behaviors, in which adaptation involves adding, removing, or reweighting components rather than updating model parameters, giving rise to two complementary lines of research.
Behavior retrieval organizes prior experiences as independently accessible records, either by retrieving trajectories~\cite{fist2022, pmlr-v205-nasiriany23a, pari2021surprising, behaviorretrieval2023, lin2024flowretrieval, 10610455, memmel2025strap, kumar2025collage, dass2026datamil, liang2026autonomous} or by routing over learned skill embeddings and prototypes~\cite{l2m2024, iscil,lotus2023, guo2025srsa, lee2025policy, lei2025dynamic}, allowing individual components to be added or removed without restructuring the entire memory.
Structured procedural representations~\cite{yin2024offline, wang2025robust, chen2026deco, shao2025symskill} instead encode explicit connectivity among reusable behavior components, supporting compositional access where a target instruction or goal is addressed by combining connected components.

However, these approaches face distinct challenges under \textit{sparse governing information}, where task goals, reusable procedural factors, and prior experiences provide fragmented evidence that does not enumerate their possible task-specific combinations. 
Under these conditions, task demonstrations provide sparse snapshots of desired state changes, while procedure records capture skill-induced transitions realizable by the underlying behavior model without specifying how to compose them for execution.
(i) Behavior retrieval typically matches stored experiences independently, leaving the transition compatibility between retrieved records under-specified.
(ii) Structured procedural approaches explicitly represent transition connectivity, but often encode it within a single state or skill abstraction, which can obscure factor-specific transition relations.
Reconfiguration of robot behavior therefore requires a structure that preserves factor-specific transition connectivity across varying contexts and supports task-conditioned composition of executable skill paths.

To address this gap, we propose \textbf{\ours{}}, a multi-view update-and-compose framework for reconfigurable skill policies that maintains editable procedure and task memories to materialize task-specific executable skill sequences. 
In \ours{}, the \textit{Procedure Knowledge Base} organizes task-agnostic skill-induced transition records into view-local graphs, preserving skill connectivity and effect evidence across reusable factors of variation. 
At test time, \textit{Task Materialization} uses these views to abstract the task demonstration into a compositional query of supported view-level changes, then resolves the query into an executable skill sequence by tracing compatible transitions across stored graphs.

We evaluate \ours{} on an extended Franka Kitchen benchmark with held-out long-horizon tasks requiring compositions beyond stored skill records. By varying skill-context diversity, view resolution, and task-demonstration waypoint gaps, we assess reconfiguration under sparse procedure and task evidence. \ours{} achieves $91.9\%$ ordered completion at $1.0$~ms of task inference per episode, outperforming retrieval and skill-chaining baselines. Real-robot studies further demonstrate deployment-time adaptation through local memory updates.

\begin{figure*}[t]
    \centering
    \vskip -0.15in
    \includegraphics[width=1.0 \linewidth]{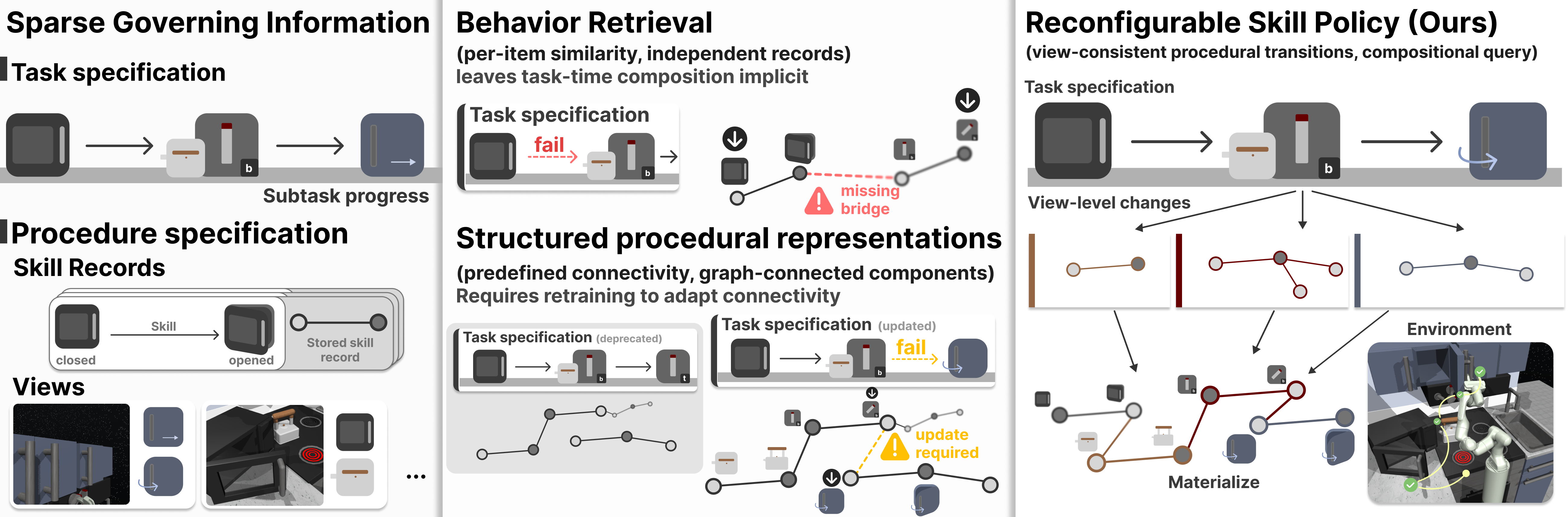}
    \label{fig:concept}
    \vskip -0.225in
\caption{
Conceptual overview of \textit{reconfigurable skill policy} under \textit{sparse governing information}.
\textit{(left)} Sparse task evidence specifies desired changes, while procedure-side evidence provides task-agnostic skill records observed through multiple procedure views.
\textit{(middle)} Behavior retrieval accesses stored records independently, whereas structured procedural approaches expose connectivity among reusable behaviors.
\textit{(right)} \ours{} converts observed view-level changes into a compositional query and materializes a task-specific executable skill sequence over view-consistent procedural transitions.
}
    \vskip -0.125in
\end{figure*}

Our contributions are as follows:
\begin{itemize}[leftmargin=*]
    \item We formulate \textit{reconfigurable skill policy} under \textit{sparse governing information} as a cost-aware update-and-compose problem over editable procedure and task memories.
    \item We propose \ours{}, a framework that composes task-agnostic skill records into task-specific executable skill sequences under a supplied procedure-view interface.
    \item We design a controlled protocol on an extended Franka Kitchen isolating procedure- and task-side sparseness, with a real-robot study of deployment-time adaptation.
\end{itemize}

\section{Related Work}
\textbf{Memory-based policy adaptation in robot learning.}
Recent robot-learning systems increasingly treat memory as an external mechanism for cost-effective policy adaptation, rather than only as hidden state or model parameters.
One line retrieves relevant demonstrations or experiences at deployment time and executes them through replay, alignment, or visual correspondence~\cite{kwon2025rt,di2024effectiveness,di2024dinobot}.
Another line uses retrieval to curate, fuse, or bootstrap task-relevant data from large-scale experience, improving few-shot generalization and robustness~\cite{fist2022,pmlr-v205-nasiriany23a,pari2021surprising, behaviorretrieval2023, lin2024flowretrieval, 10610455, memmel2025strap,kumar2025collage,dass2026datamil,liang2026autonomous}.
Recent robot policies further connect memory with in-context adaptation, retrieving task-relevant keyframes or sub-trajectories as context for VLA controllers and decision-transformer policies~\cite{schmied2025retrievalaugmented, lee2023incontext,sridhar2025memer}.
These in-context uses bring memory closer to policy execution, but still leave procedural recomposition inside the retrieved context or policy behavior rather than exposing it as a memory operation.
This makes changes in object bindings, goal orders, contexts, or skill granularities difficult to update, inspect, and measure at test time.

\textbf{Routing-based policy reconfiguration.}
In lifelong learning, such retrieval capabilities integrate external memory into learning and decision-making, separating generalized skill knowledge in pretrained policies from task-specific composition delegated to test-time routing~\cite{mai2026survey}. These approaches treat memory as support for selecting behaviors, routing over an expandable set of expert modules via similarity matching~\cite{l2m2024, iscil, lotus2023, guo2025srsa}, merging retrieved skills into connected behaviors~\cite{lei2025dynamic}, or matching trajectory distributions at runtime to reconcile modules trained at different stages~\cite{lee2025policy}.
However, collapsing diverse factors of variation into a single embedding space coarsens the state abstraction, and routing it from sparse governing information limits forward transfer~\cite{abel2018state, fitzgerald2021abstraction}. This motivates decomposing behavior memory along partial views with local connectivity within each view.

\textbf{Structured procedural representations.}
A complementary line exposes connectivity among reusable behaviors explicitly. Rooted in the options framework~\cite{sutton1999between}, these methods chain temporally extended primitives through their initiation and termination conditions to form explicit procedural structure~\cite{konidaris2009skill, Bagaria2020Option, bagaria2021skill}. A parallel direction composes options through learned models or weighted combinations of primitives~\cite{silver2012optioncomp, barreto2019option, alegre2025constructing}. Building on this connectivity view, recent works organize behavior memory as graphs over states, transitions, or skills and search them for temporal connectivity, including deep skill graphs over discovered options~\cite{bagaria2021skill, bagaria2025intrinsically}, graph-based retrieval and stitching for goal-conditioned imitation~\cite{yin2024offline, wang2025robust}, and skill composition over learned operators~\cite{chen2026deco, shao2025symskill}.
However, this connectivity is built along a single axis, typically temporal adjacency, and is fused with its underlying representation, leaving no handle to reorganize the same records under sparse task evidence. We propose \ours{}, which stores skill-induced records under multiple partial views with local connectivity and materializes an executable structure across views when a task arrives.

\section{Preliminaries}
\textbf{Controlled semi-Markov process~\cite{sutton1999between}.}
We model the environment as a \textit{controlled Markov process} (CMP) $\mathcal{M} = (\mathcal{S}, \mathcal{A}, \mathcal{P}_{\mathcal{A}})$ with state space $\mathcal{S}$, primitive action space $\mathcal{A}$, and one-step transition kernel $\mathcal{P}_{\mathcal{A}}(s' \mid s, a)$. Given a set of skills $\Omega$, where each $\omega \in \Omega$ is a temporally extended controller over $\mathcal{A}$, the environment induces a \textit{controlled semi-Markov process} (CSMP) $\mathcal{M}_{\Omega} = (\mathcal{S}, \Omega, \mathcal{P}_{\Omega})$, with semi-Markov transition kernel $\mathcal{P}_{\Omega}(s', k \mid s, \omega)$ over termination state $s' \in \mathcal{S}$ and duration $k \in \mathbb{N}_{+}$.

\textbf{SMDP homomorphism~\cite{ravindran2003smdp, ravindran2004algebraic, rezaei2022continuous}.}
For a given task $\tau$, equipping the CSMP $\mathcal{M}_{\Omega}$ with a task-dependent reward
$\mathcal{R}_{\tau} : \mathcal{S} \times \Omega \times \mathbb{N}_{+} \to \mathbb{R}$,
where $\mathcal{R}_{\tau}(s,\omega,k)$ denotes the reward of executing skill $\omega$ from state $s$ for duration $k$,
yields the task-specific SMDP
$\mathcal{M}_{\Omega,\tau} = (\mathcal{S}, \Omega, \mathcal{P}_{\Omega}, \mathcal{R}_{\tau})$.
A task-preserving abstraction of $\mathcal{M}_{\Omega,\tau}$ is specified by an SMDP homomorphism $h_{\tau} = (f_{\tau}, \{g_{\tau,s}\}_{s \in \mathcal{S}})$ from $\mathcal{M}_{\Omega,\tau}$ to an abstract SMDP $\bar{\mathcal{M}}_{\Omega,\tau}$,
where $f_{\tau} : \mathcal{S} \to \bar{\mathcal{S}}_{\tau}$ is a surjective state map and
$g_{\tau,s} : \Omega \to \bar{\Omega}_{\tau}$ is a state-dependent skill map.
The map $h_{\tau}$ is an SMDP homomorphism if concrete state-skill pairs mapped to the same
abstract state-skill pair induce identical aggregate semi-Markov transition kernels and task rewards.
Specifically, for any $(s_1,\omega_1),(s_2,\omega_2) \in \mathcal{S}\times\Omega$
satisfying $f_{\tau}(s_{1}) = f_{\tau}(s_{2})$ and
$g_{\tau,s_{1}}(\omega_{1}) = g_{\tau,s_{2}}(\omega_{2})$,
the following condition holds for all $\bar{s}' \in \bar{\mathcal{S}}_{\tau}$ and
$k \in \mathbb{N}_{+}$:
\begin{equation}
\sum_{s' : f_\tau(s') = \bar{s}'} \!\!\!\!\!\! \mathcal{P}_{\Omega}(s', k \mid s_1, \omega_1)
=
\sum_{s' : f_\tau(s') = \bar{s}'} \!\!\!\!\!\! \mathcal{P}_{\Omega}(s', k \mid s_2, \omega_2),
\quad
\mathcal{R}_{\tau}(s_1,\omega_1,k)=\mathcal{R}_{\tau}(s_2,\omega_2,k).
\label{eq:smdp_hom}
\end{equation}
These conditions induce the abstract SMDP
$\bar{\mathcal{M}}_{\Omega,\tau}
= (\bar{\mathcal{S}}_{\tau}, \bar{\Omega}_{\tau}, \bar{\mathcal{P}}_{\tau}, \bar{\mathcal{R}}_{\tau})$.

\section{Problem Formulation}
We formalize a reconfigurable skill policy as an `update-and-compose' rule over editable procedure and task memories, producing task-specific executable realizations via task-local abstractions.

\textbf{Reconfigurable skill policy.}
Let $p \in \mathbb{N}_{+}$ index discrete phases, with superscript $(\cdot)^{p}$ indicating phase $p$. When no ambiguity arises, quantities without superscripts denote their current phase-$p$ values.
At each phase, two specifications $(\varphi_{\Omega}^{p}, \varphi_{\tau}^{p})$ are given. The procedure-side specification $\varphi_{\Omega}^{p}$ supplies evidence and structural information about the currently realizable skills $\Omega$, and the task-side specification $\varphi_{\tau}^{p}$ supplies task-state evidence and goal-side updates for the task $\tau$ under consideration~\cite{jothimurugan2019composable}.
The reconfigurable skill policy $\pi_{\Omega,\tau}^p$ maintains a paired procedure and task memory $(\mathcal{K}_{\Omega}^{p}, \mathcal{K}_{\tau}^{p})$ and a composition map $\phi$. The phase specifications update the memory pair, and the updated pair is composed as follows:
\begin{equation}
\left(
\mathcal{K}_{\Omega}^{p},
\mathcal{K}_{\tau}^{p}
\right)
=
\left(
U_{\Omega} (\mathcal{K}_{\Omega}^{p-1}, \varphi_{\Omega}^{p}),
U_{\mathcal{T}}\!\left(\mathcal{K}_{\tau}^{p-1}, \varphi_{\tau}^{p}\right)
\right),
\qquad
\hat{h}_{\tau}^{p}
=
\phi\!\left(\mathcal{K}_{\Omega}^{p}, \mathcal{K}_{\tau}^{p}, \tau\right).
\label{eq:memory_update}
\end{equation}
Accordingly, the procedure update $U_{\Omega}$ instantiates and revises the currently realizable
set of skill $\Omega$ and its induced phase-local CSMP $\mathcal{M}_{\Omega}^{p}$,
while the task update $U_{\mathcal{T}}$ revises the task-side memory and the specification associated with $\tau$.
The composition map $\phi$ maps the updated memory pair and task $\tau$ to
$\hat{h}_{\tau}^{p}$, which induces an abstract SMDP
$\bar{\mathcal{M}}_{\Omega,\tau}^{p}$ by \eqref{eq:smdp_hom}.

In practice, $\phi$ is realized as a composition of partial
homomorphisms~\cite{ravindran2004algebraic}, so that $\hat{h}_{\tau}^{p}$ and
$\bar{\mathcal{M}}_{\Omega,\tau}^{p}$ are approximations of an SMDP
homomorphism and its induced abstract SMDP rather than exact ones.
At execution time, an abstract decision
$\bar{\pi}_{\tau}^{p}(\bar{\omega} \mid f_{\tau}^{p}(s))$
is grounded into a concrete skill through a task-conditioned realization rule
that selects an executable $\omega \in \Omega$ consistent with the abstract
skill $\bar{\omega}$ and the current state $s$, yielding the reconfigurable
skill policy $\pi_{\Omega,\tau}^{p}(\omega \mid s)$.

\textbf{Objective.}
Let $\mathcal{T}^{p}$ denote the set of tasks considered at phase $p$. The
procedure-side specification $\varphi_{\Omega}^{p}$ is shared across
$\mathcal{T}^{p}$, while each $\tau \in \mathcal{T}^{p}$ is associated with
its own task-side specification $\varphi_{\tau}^{p}$.
Under changing environments and skill interfaces, the objective is to find
update-and-compose operators $(U_{\Omega}^{*}, U_{\mathcal{T}}^{*}, \phi^{*})$
that maximize per-task return while accounting for materialization and update
costs:
\begin{equation}
(U_{\Omega}^{*}, U_{\mathcal{T}}^{*}, \phi^{*})
\in
\underset{U_{\Omega},U_{\mathcal{T}},\phi}{\arg\max}
\sum_{p}
\left[
\sum_{\tau \in \mathcal{T}^{p}}
\left(
J_{\tau}(\pi_{\Omega,\tau}^{p};\mathcal{M}_{\Omega,\tau}^{p})
-
\mathcal{C}_{\phi}^{p}(\tau)
\right)
-
\mathcal{C}_{U}^{p}
\right].
\label{eq:reconfigurable_policy_objective}
\end{equation}
Here, the return $J_{\tau}(\pi_{\Omega,\tau}^{p};\mathcal{M}_{\Omega,\tau}^{p})$
is evaluated on the phase-$p$ task SMDP, with $\pi_{\Omega,\tau}^{p}$ defined
by \eqref{eq:memory_update}. The task-wise cost
$\mathcal{C}_{\phi}^{p}(\tau)$ accounts for applying $\phi$ to materialize the
task-specific abstraction, while the phase-wise cost $\mathcal{C}_{U}^{p}$
accounts for updating the procedure and task-side memories at phase $p$.

\section{Methods}

Following the update-and-compose formulation, \ours{} instantiates \eqref{eq:memory_update} through two named memory interfaces that separate procedure storage from task-conditioned composition. 
Across these interfaces, multiple procedure views preserve distinct state equivalences over the same skill-induced transitions, allowing task materialization to combine task-relevant distinctions into a task-specific abstraction.

\textbf{(i) Procedure knowledge base.}
The \textit{Procedure Knowledge Base} realizes the procedure-memory pair $(\mathcal{K}_{\Omega}^{p},U_{\Omega})$. It updates and stores skill-induced transition evidence as procedure records, reads them through partial procedure
views, and organizes the resulting view-projected records into skill-transition graphs and correlation-marked records.
This yields a persistent procedure substrate whose reusable executable transitions support later reconfiguration.

\begin{figure*}[t]
    \centering
    \includegraphics[width=1.0 \linewidth]{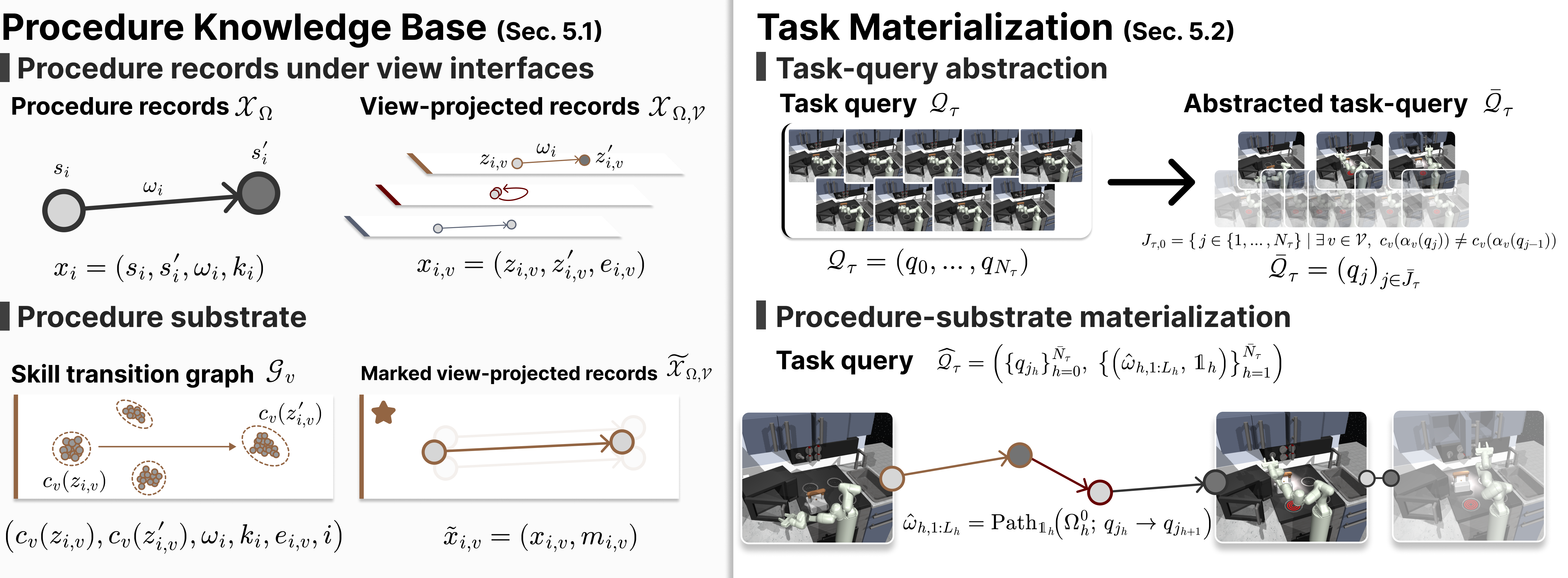}
    \vskip -0.1in
    \caption{
    \ours{} realizes update-and-compose through two memory interfaces.
    \textit{(left)} The \textit{Procedure Knowledge Base} turns procedure records and view interfaces into a persistent procedure substrate of skill-transition graphs and salience marks.
    \textit{(right)} \textit{Task Materialization} turns a task query into an abstracted query over the stored substrate and materializes it as an executable skill path.
    }
    \label{fig:method_overview}
    \vskip -0.1in
\end{figure*}

\textbf{(ii) Task materialization.}
The task-memory pair $(\mathcal{K}_{\tau}^{p},U_{\mathcal{T}})$ maintains current task-state evidence. The composition map $\phi$ is realized by \textit{Task Materialization}, which folds this evidence into the same canonical view-node space, forms task-query abstractions, and completes the induced query over the procedure substrate as current-state-conditioned skill paths.
The resulting executable approximation of the task quotient carries prefix states and masks for grounding and local rematerialization during execution. Figure~\ref{fig:method_overview} provides an overview.

\subsection{Procedure Knowledge Base}
\textbf{Procedure records under view interfaces.}
At phase $p$, the procedure-side evidence $\varphi^p_\Omega$ provides observed
executions under the current skills $\Omega$.
Let $\mathcal{I}_p$ with $|\mathcal{I}_p|=N_p$ denote the globally unique
identifiers assigned to records newly added at phase $p$.
\ours{} updates the procedure-record set $\mathcal{X}^p_\Omega$
by appending a record $x_i$ for each $i\in\mathcal{I}_p$:
\begin{equation}
    \mathcal{X}^p_\Omega
    =
    \mathcal{X}^{p-1}_\Omega
    \cup
    \{x_i\}_{i\in\mathcal{I}_p},
    \qquad
    x_i=(s_i,s'_i,\omega_i,k_i).
    \label{eq:procedure-record}
\end{equation}
Here, $s_i,s'_i\in\mathcal{S}$ are segment endpoints, $\omega_i\in\Omega$
is the executed skill, and $k_i\in\mathbb{N}_{+}$ is its duration. Each record
preserves executable transition evidence before it is organized for
task-specific materialization.

Since the same record may be reused along different aspects of its
transition under different tasks, \ours{} reads each record through
partial procedure views $\mathcal{V}$, with each view preserving one
such aspect for task materialization.
The view set $\mathcal{V}$ is supplied as part of the procedure-side specification, either specified directly or induced from state factorization, and each view $v\in\mathcal{V}$ provides a \textit{view interface} consisting of a projection $\alpha_v:\mathcal{S}\rightarrow\mathcal{Z}_v$ and a view-local effect map $\delta_v$ on $\mathcal{Z}_v$.
Pairing each record $x_i\in\mathcal{X}_\Omega$ with each $v\in\mathcal{V}$ yields the view-projected record collection $\mathcal{X}_{\Omega,\mathcal{V}}$,
\begin{equation}
    \mathcal{X}_{\Omega,\mathcal{V}}
    =
    \{x_{i,v}\mid x_i\in\mathcal{X}_\Omega,\; v\in\mathcal{V}\},
    \qquad
    x_{i,v}=(z_{i,v},z'_{i,v},e_{i,v}).
    \label{eq:view-projected-records}
\end{equation}
Here, $x_{i,v}$ is the view-projected record obtained by reading $x_i$ through
view $v$, with projected start $z_{i,v}=\alpha_v(s_i)$, projected endpoint
$z'_{i,v}=\alpha_v(s'_i)$, and view-local effect
$e_{i,v}=\delta_v(z_{i,v},z'_{i,v})$.

\textbf{Procedure substrate.}
To expose local connectivity among projected procedure effects, \ours{}
canonicalizes the view-projected records into per-view skill-transition graphs.
For each view $v\in\mathcal{V}$, data-adaptive clustering of projected record
endpoints (\textit{e.g.}, DBSCAN~\cite{dbscan}) induces canonical
view nodes $\mathcal{N}_v$ and a canonicalization map
$c_v:\mathcal{Z}_v\rightarrow\mathcal{N}_v$. Each view-projected record
$x_{i,v}$ then contributes a directed edge between its canonicalized
endpoints, forming the view-level skill-transition graph $\mathcal{G}_v$:
\begin{equation}
\mathcal{G}_{v}
=
(\mathcal{N}_{v},\mathcal{E}_{v}),
\qquad
\mathcal{E}_{v}
=
\left\{
\left(
c_v(z_{i,v}),
c_v(z'_{i,v}),
\omega_i,
k_i,
e_{i,v},
i
\right)
\mid
x_i\in\mathcal{X}_\Omega
\right\}.
\label{eq:view-skill-transition-graph}
\end{equation}
Here, each edge in $\mathcal{E}_v$ runs from the canonicalized start
$c_v(z_{i,v})$ to the canonicalized endpoint $c_v(z'_{i,v})$, carrying the
executed skill $\omega_i$, duration $k_i$, view-local effect $e_{i,v}$, and
procedure-record pointer $i$. The collection
$\mathcal{G}_{\Omega,\mathcal{V}}=\{\mathcal{G}_{v}\}_{v\in\mathcal{V}}$
forms the view-level skill-transition graph collection.

The same view-projected records are also annotated with salience evidence.
For each $x_{i,v}$, \ours{} computes a salience mark
$m_{i,v}=\Gamma(x_{i,v};\mathcal{X}_{\Omega,\mathcal{V}})$, which measures
local skill-label consistency among neighboring view-local effects.
This induces the marked view-projected records
collection $\bar{\mathcal{X}}_{\Omega,\mathcal{V}}$:
\begin{equation}
\bar{\mathcal{X}}_{\Omega,\mathcal{V}}
=
\{
\bar{x}_{i,v}
\mid
x_{i,v}\in\mathcal{X}_{\Omega,\mathcal{V}}
\},
\qquad
\bar{x}_{i,v}
=
(x_{i,v},m_{i,v}).
\label{eq:correlation-marked-records}
\end{equation}
Together, the pair $(\mathcal{G}_{\Omega,\mathcal{V}},\,
\bar{\mathcal{X}}_{\Omega,\mathcal{V}})$ defines the \textit{procedure
substrate} stored in $\mathcal{K}_{\Omega}$.

\subsection{Task Materialization}
\label{sec:task-query-abstraction}
To realize $\pi_{\Omega,\tau}^{p}$, \ours{} instantiates the composition map
$\phi$ in two stages over the task query
$\mathcal{Q}_{\tau}=(q_{0},\ldots,q_{N_{\tau}})$ from $\mathcal{K}_{\tau}$,
a sequence of task-side waypoints. The first stage abstracts
$\mathcal{Q}_{\tau}$ into a query against the precomputed procedure substrate
$(\mathcal{G}_{\Omega,\mathcal{V}},\bar{\mathcal{X}}_{\Omega,\mathcal{V}})$,
and the second materializes the abstracted query as executable skill bindings.

\textbf{Task-query abstraction.}
To form the initial query $\mathcal{Q}_{\tau,0}$, \ours{} projects each $q_{j}$ into its per-view node assignment $\{c_{v}(\alpha_{v}(q_{j}))\}_{v\in\mathcal{V}}$ and collects \textit{decision points}, the steps at which at least one view node differs from the preceding step, into $\mathcal{Q}_{\tau,0}$,
\begin{equation}
    \mathcal{Q}_{\tau,0}
    =
    (q_{j})_{j \in J_{\tau,0}},
    \qquad
    J_{\tau,0}
    =
    \{\,j \in \{1,\ldots,N_{\tau}\} \mid \exists\, v \in \mathcal{V},\ c_{v}(\alpha_{v}(q_{j})) \neq c_{v}(\alpha_{v}(q_{j-1}))\,\}.
    \label{eq:initial-query}
\end{equation}
Here, $\mathcal{Q}_{\tau,0}$ is the subsequence of $\mathcal{Q}_{\tau}$ at the decision points indexed by $J_{\tau,0}$.

To refine the initial query $\mathcal{Q}_{\tau,0}$ into a substrate-supported subsequence, \ours{} traverses $\mathcal{Q}_{\tau}$ step by step.
From the current decision point, $\mathcal{G}_{\Omega,\mathcal{V}}$ exposes
which view-node transitions are reachable through stored records, while
$\bar{\mathcal{X}}_{\Omega,\mathcal{V}}$ supplies the marked transition
evidence used to score whether the observed task displacement agrees with
stored view-local effects.
The next abstracted decision point is selected from candidate steps containing
a changed view in which some stored skill attains its highest mean salience mark
$m_{i,v}$ over its records.
Iterating along $\mathcal{Q}_{\tau}$ and retaining the decision points that stored records support as segment boundaries yields the abstracted query $\bar{\mathcal{Q}}_{\tau}$,
\begin{equation}
    \bar{\mathcal{Q}}_{\tau}
    =
    (q_{j})_{j \in \bar{J}_{\tau}},
    \qquad
    \bar{J}_{\tau}
    =
    \{j_{0}, j_{1}, \ldots, j_{\bar{N}_{\tau}}\} \subseteq \{0\} \cup J_{\tau,0},
    \qquad
    0 = j_{0} < \cdots < j_{\bar{N}_{\tau}} \leq N_{\tau}.
    \label{eq:abstracted-query}
\end{equation}
Here, $\bar{\mathcal{Q}}_{\tau}$ collects the decision points of $\mathcal{Q}_{\tau}$ supported by the procedure substrate, indexed by $\bar{J}_{\tau}$ with start index $0$ prepended, partitioned into $\bar{N}_{\tau}$ segments for executable binding.

\textbf{Procedure-substrate materialization.}
Each consecutive pair $(q_{j_{h}}, q_{j_{h+1}})$ in $\bar{\mathcal{Q}}_{\tau}$
specifies a local segment with active views $\mathcal{V}_{h}\subseteq\mathcal{V}$,
the views in which the canonical node changes, and a desired view-local
effect on each record, with the indicator $\mathds{1}_{h} = \mathds{1}[\mathcal{V}_{h}]$
used at execution time for grounding.
Restricting the marked records $\bar{\mathcal{X}}_{\Omega,\mathcal{V}}$
to $\mathds{1}_{h}$ yields the admissible skill set
$\Omega_{h}\subseteq\Omega$ with stored view-local effects on
the active views of the segment.

For each skill in $\Omega_{h}$, \ours{} selects the view in which the
records of the skill attain the highest mean salience mark. In this view, the
stored view-local effect most reliably distinguishes the skill from other stored skills.
\ours{} keeps a skill when, in this view, its typical stored effect points in
the direction of the change that the segment demands.
Because each skill is compared only in its own view, skills that account for
different parts of this change are kept together in the candidate set
$\Omega^{0}_{h}\subseteq\Omega_{h}$.

\ours{} then binds each segment by applying the path operator to the
candidate skill set under $\mathds{1}_{h}$,
\begin{equation}
    \hat{\omega}_{h,1:L_{h}}
    =
    \mathrm{Path}_{\mathds{1}_{h}}\!\bigl(\Omega^{0}_{h};\, q_{j_{h}} \to q_{j_{h+1}}\bigr),
    \qquad
    L_{h} = |\Omega^{0}_{h}|.
    \label{eq:segment-binding}
\end{equation}
Here, $\mathrm{Path}_{\mathds{1}_{h}}$ returns a chained executable skill
sequence over $\Omega_h^0$ between the segment endpoints, ordering all
selected skills so that each starts near where the previous one ends on the
active views. The candidate
selection rule and path-search objective are specified in
Appendix~\ref{app:task_materialization_details}.

Finally, the segment bindings form the materialized object
\begin{equation}
    \widehat{\mathcal{Q}}_{\tau}
    =
    \Bigl(
      \{q_{j_{h}}\}_{h=0}^{\bar{N}_{\tau}},\;
      \bigl\{\bigl(\hat{\omega}_{h,1:L_{h}},\, \mathds{1}_{h}\bigr)\bigr\}_{h=0}^{\bar{N}_{\tau}-1}
    \Bigr)
    \label{eq:materialized-task}
\end{equation}
is an executable approximation of the task quotient. The decision
points $\{q_{j_h}\}$ carry the abstract decisions, and each segment
$h$ supplies an executable skill sequence $\hat{\omega}_{h,1:L_{h}}$
together with its active-view indicator $\mathds{1}_{h}$ used for
grounding. At execution, the runtime advances to segment $h{+}1$
only when the view-node transition under $\mathds{1}_{h}$ is observed,
and a local inconsistency triggers rematerialization of the affected
segment alone.

\section{Experiments}
Our experiments evaluate whether \ours{} can reconfigure policies under \textit{sparse governing information}, manifesting in procedure specifications that under-determine how to compose stored skills, and in task specifications that under-determine which composition to produce. We additionally examine whether these gains transfer to a real robot. We structure the evaluation around three questions:

\noindent\textbf{RQ1.} Is \ours{} robust across varying \textit{skill-context diversity} and \textit{view resolution}?

\noindent\textbf{RQ2.} Can \ours{} recover executable composition under sparse task specifications?

\noindent\textbf{RQ3.} Can \ours{} adapt a real-world robot through local procedure or task updates?

\begin{figure}[t]
    \centering
    \includegraphics[width=.96\linewidth]{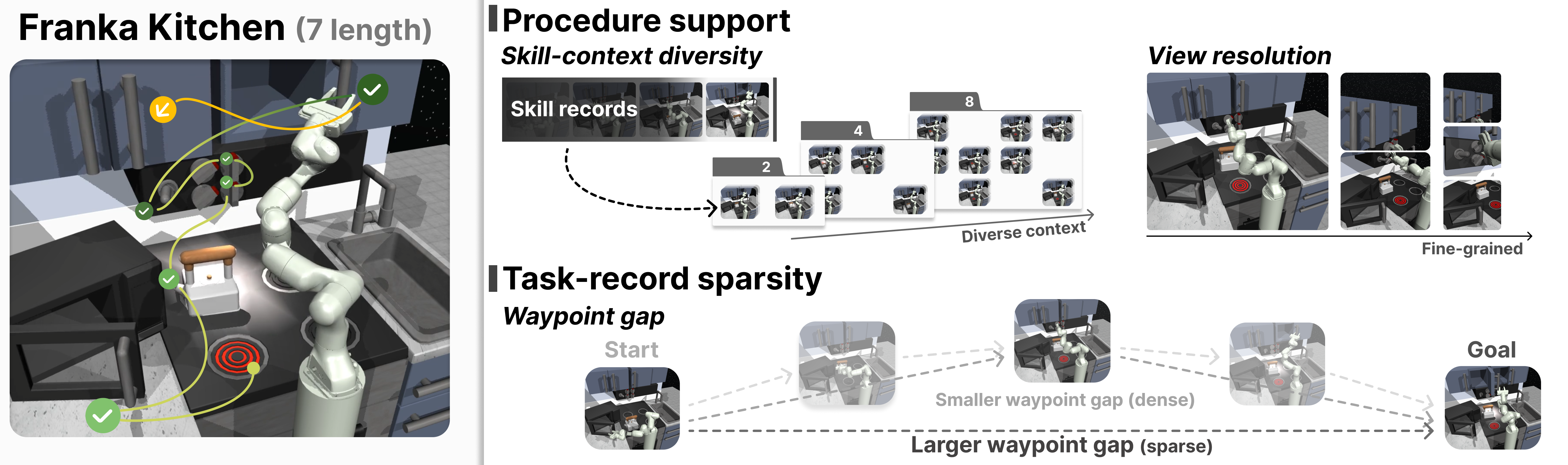}
    \vskip -0.05in
\caption{\textit{(left)} Franka Kitchen with an example length-$7$ target task, held out from the procedure knowledge base.
\textit{(right)} Evaluation axes.
\textit{Procedure support} varies \textit{skill-context diversity}, the number of length-$4$ task realizations populating the procedure knowledge base, and \textit{view resolution}, the granularity of view projections supplied at inference.
\textit{Task-record sparsity} varies the \textit{waypoint gap} $H$, the stride in subtask indices between consecutive task-record waypoints.}
\vskip -0.1in
    \label{fig:environments}
\end{figure}

\subsection{Experiment Settings}

\textbf{Environment.} We build on Franka Kitchen~\cite{gupta2019relay, fu2020d4rl} by extending it with a broader set of causally independent~\cite{caiac2024} subtasks that admit varied executions, expanding the space of combinatorial long-horizon compositions and the diversity of contexts in which each skill is observed. We define realization length as the number of subtasks executed in sequence, and extend the original $24$ length-$4$ realizations of Franka Kitchen up to length-$7$.

\textbf{Evaluation protocol.} We design the evaluation to stress-test the update-and-compose setting under sparse governing information by varying \textit{procedure support} and \textit{task-record sparsity}. The evaluation split uses length-$4$ realizations to construct the procedure knowledge base and introduces a held-out length-$7$ target task at test time whose complete skill composition is absent from the stored records.

\textit{Procedure support} stresses the \textit{procedure knowledge base} by sweeping \textit{skill-context diversity}, the variety of contexts in which each skill is observed, over $\{4, 8, 16, 32\}$ length-$4$ subtask orderings populating the procedure knowledge base.
For the view axis, the procedure-view interface is built from Kitchen-defined object-factor views and a robot-state view, with the finest setting separating them into up to eight atomic views.
\textit{View resolution} varies the number of available views over $\{3, 4, 5, 8\}$ by randomly merging object factors into composite views.
\textit{Task-record sparsity} stresses \textit{Task Materialization} by sweeping the \textit{waypoint gap} $H \in \{1, ... , 7\}$, the stride between consecutive task-record waypoints in subtask indices, with the final goal appended whenever the stride does not reach it. For $H{=}4$, this yields the waypoint list \texttt{[0,4,7]}.
For each experimental configuration, we use $10$ random seeds and $100$ evaluation episodes per seed. We report \textit{ordered completion} and per-episode wall-clock task inference time.
Figure~\ref{fig:environments} provides an overview, with details in Appendix~\ref{app:evaluation-protocol}.

\textbf{Baselines.} We evaluate all baselines under a shared reconfiguration setting that varies \textit{procedure support} and \textit{task-record sparsity} to assess how different composition mechanisms operate under \textit{sparse governing information}. 
\texttt{Skill Retrieval} (\texttt{SR}) first computes a multi-view score for each candidate skill by matching each waypoint transition to the closest stored record for that skill in each procedure view, then independently reranks skills by the averaged view-wise score.
\texttt{Skill Sequence Retrieval} (\texttt{SSR}) extends this retrieval to the combinatorial space of per-step skill candidates, ranking candidate sequences by aggregated retrieval scores without explicitly modeling inter-skill connectivity.
\texttt{Skill Chaining} (\texttt{SC})~\cite{konidaris2009skill, bagaria2021skill} represents procedural connectivity through a skill-level transition graph. Each skill is summarized by canonical initiation and termination states estimated from stored endpoints, with graph edges weighted by termination-to-initiation distances.

\textbf{Metrics.} We report \textit{ordered completion}, the fraction of environment-defined subtasks completed in the demonstrated order~\cite{mees2022calvin}, and \textit{task inference time}, the total time spent inferring an executable skill sequence per episode.

\begin{figure}[h!]
    \centering
    \includegraphics[width=\linewidth]{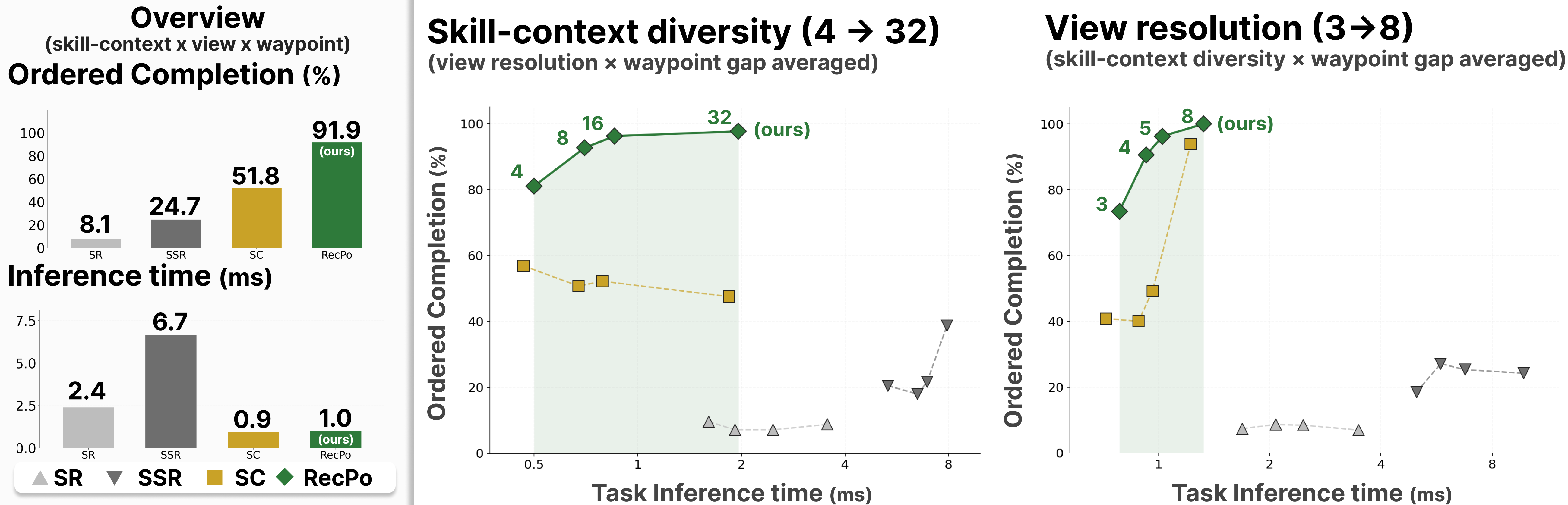}
    \vskip -0.05in
\caption{
\textit{Ordered completion}--task inference time trade-off under varying \textit{procedure support} and \textit{task-record sparsity}.
\textit{(left)} Overall comparison of the three baselines and \ours{}, averaged across all evaluation conditions.
\textit{(right)} Trade-offs under varying \textit{procedure support}, sweeping \textit{skill-context diversity} and \textit{view resolution} separately while averaging over the remaining axes.
}
    \label{fig:reconfig}
\end{figure}

\begin{figure}[h]
     \centering
    \vskip -0.05in
     \includegraphics[width=1\linewidth]{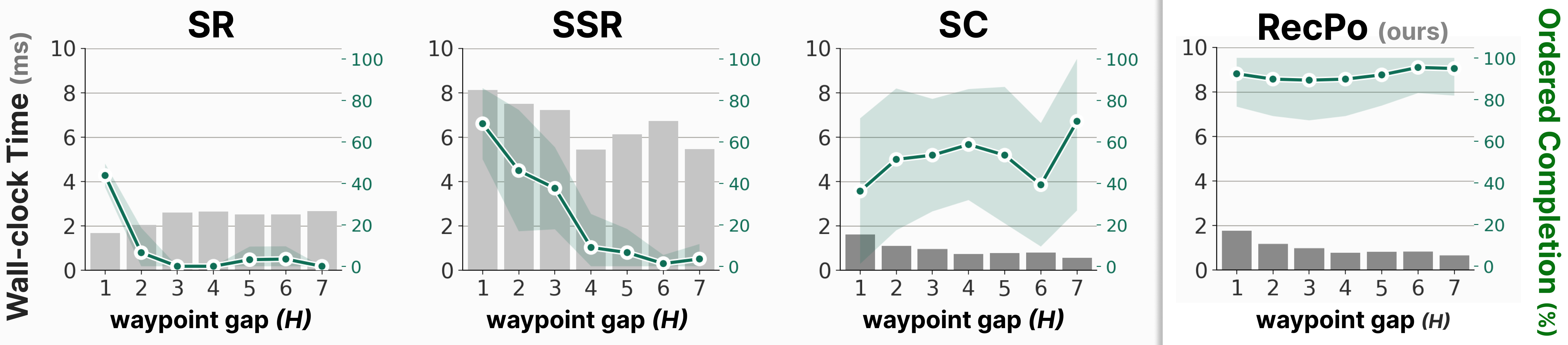}
    \vskip -0.1in
\caption{
\textit{Ordered completion} and task inference time under varying \textit{task-record sparsity}.
Results for the three baselines and \ours{} are reported across \textit{waypoint gap} $H$, averaged over \textit{skill-context diversity} and \textit{view resolution}.
Gray bars indicate per-episode task inference time (ms, left axis), while green lines indicate \textit{ordered completion} (\%, right axis).
Shaded bands indicate one standard deviation of \textit{ordered completion} across procedure-support configurations at each $H$.
}
    \vskip -0.15in
     \label{fig:main2}
\end{figure}

\subsection{Procedure Knowledge Base}
To answer \textbf{RQ1}, we evaluate \ours{} and all baselines under sparse procedure specifications by varying \textit{skill-context diversity} and \textit{view resolution}, with smaller values of either instantiating sparser regimes.
Figure~\ref{fig:reconfig}\,\textit{(left)} shows that \ours{}, which narrows candidates to view-consistent transitions, achieves the highest \textit{ordered completion} at $91.9\%$ with only $1.0$~ms of task inference per episode. \texttt{SSR} widens the search to the combinatorial space of skill sequences but reaches $24.7\%$ at $6.7$~ms, suggesting that naive scaling of the search space offers limited gains without structural constraints on composition.

Figure~\ref{fig:reconfig}\,\textit{(right)} shows that task inference time increases for every method with higher \textit{skill-context diversity} or \textit{view resolution}, while every \ours{} configuration remains on the Pareto front of completion and time.
\texttt{SC} declines from $56.9\%$ to $47.5\%$ as skill-context diversity increases and approaches \ours{} only at the highest view resolution, since the skill-level graph of \texttt{SC} aliases endpoints across contexts, and higher view resolution still leaves hard negatives whose endpoints match but whose effects do not.
\ours{} instead structures salience marks with skill-transition graphs, so records remain distinguishable by the views they govern, keeping composition robust under sparse \textit{procedure support}.

\subsection{Task Materialization}
To answer \textbf{RQ2}, we evaluate \ours{} and all baselines under sparse task specifications by varying the \textit{waypoint gap} $H \in \{1, \ldots, 7\}$ over the length-$7$ target task.
Figure~\ref{fig:main2} shows that \ours{} maintains $89.3$--$95.4\%$ \textit{ordered completion} across all $H$ while its task inference time decreases from $1.8$ to $0.7$~ms.
\texttt{SR} and \texttt{SSR} collapse as $H$ increases, with \texttt{SSR} falling from $68.7\%$ to at most $9.1\%$ for $H \geq 4$, since similarity alone surfaces stored segments that do not realize demanded subtasks absent from the procedure knowledge base.
Unlike retrieval, \texttt{SC} and \ours{} do not degrade with sparser waypoints, since each waypoint fixes a segment to bind, and denser waypoints multiply the bindings over which errors accumulate.
\texttt{SC} accumulates the aliasing identified in RQ1 across these bindings and ranges from $36.2\%$ to $69.9\%$, whereas \ours{} binds each segment to view-consistent transitions and keeps the accumulated error small.

\begin{figure}[t]
     \centering
     \includegraphics[width=1\linewidth]{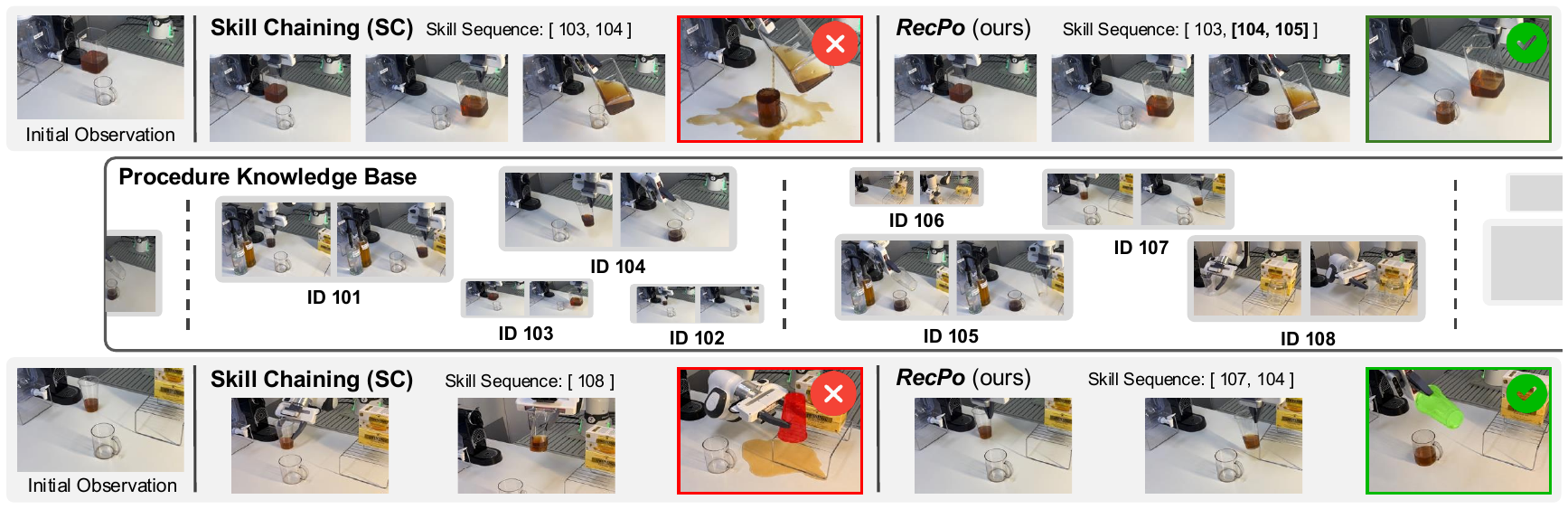}
     \vskip -0.05in
    \caption{Real-world use cases with the shared goal \textit{pour coffee}. Records 101 to 104 alone leave both cases unsolved, so the procedure knowledge base is locally updated with records 105 to 108 and a view for the target fill level, over which \texttt{SC} and \ours{} compose. Bracketed record IDs denote one segment materialized as a multi-skill path.
\textit{(top)} The source holds more coffee than the target cup can take, so pouring must stop before the cup fills.
\textit{(bottom)} The scene shares its rack and box background with records 106 to 108, while pouring record 104 was recorded on a plain table.}
     \vskip -0.1in
     \label{fig:realworld}
\end{figure}

\subsection{Use Cases} %
To answer \textbf{RQ3}, we test on a real Franka robot whether local updates adapt \ours{} along two axes without retraining, where view updates change what composition distinguishes and record updates change what it can bind, in the two cases of Figure~\ref{fig:realworld}.

In the first case of Figure~\ref{fig:realworld}\,\textit{(top)}, \texttt{SC} chains pouring record 104 alone and overflows the cup, reflecting the aliasing observed in RQ1. Even with the fill-level view added, its endpoint match still weighs the other views without salience, so stopping record 105, recorded in a different background, does not enter the chain. \ours{} weighs record 105 by its salience on the fill-level view and rematerializes the pouring segment as a skill path chaining records 104 and 105.

In the second case of Figure~\ref{fig:realworld}\,\textit{(bottom)}, \texttt{SC} binds record 108, a hard negative that also pours but inverts the cup, since its endpoints match the shared background while its effect does not realize the demanded fill-level transition, and spills the contents. \ours{} binds pouring record 104 by its effect on the fill-level view after record 107, so the background difference does not mislead the binding.

\section{Conclusion}
We present \ours{}, which formulates the reconfiguration of skill policies as test-time composition over an editable and inspectable procedure-memory interface. \ours{} stores skill-induced transition evidence as task-agnostic records and organizes them through procedure views. From sparse task evidence, \ours{} materializes task-specific executable skill sequences without retraining. Results show that this structured materialization improves the success-cost trade-off of reconfiguration across sparse procedure support and sparse task specifications, supporting adaptation through recomposition.

\textbf{Limitations and future directions.}
Our work investigates the reconfiguration of procedural knowledge through task-conditioned composition over supplied procedure views. This formulation assumes the availability of views that expose task-relevant skill effects and their connectivity. Inadequate view decomposition may obscure these relations and limit reconfigurability. 
These views can still be separated counterfactually to match the intent of the user without retraining, and extending this interface beyond object-centric factors and context generalization remains open.
Beyond the views, each skill record keeps only its start and end states, which suffices for composing skills whose effects transfer across contexts but not for skills whose outcome depends on how they are executed, and editable records that also carry such execution conditions are a natural extension.

As a future direction, the editability and inspectability of the memory interface can be leveraged for reconfiguration across diverse scenarios, such as interactive updates on resource-limited embedded devices and guardrails, since composition realizes only transitions supported by the memory and editing its views and records therefore bounds what a robot can do.

\newpage
\section*{Acknowledgment}
This work was supported
by the Institute of Information \& Communications Technology Planning \& Evaluation (IITP) grant funded by the Korea
government (MSIT)
(RS-2025-02218768, Accelerated Insight Reasoning via Continual Learning (50\%);
RS-2026-25505492, High-Performance Research AI Computing Infrastructure Support at the 2 PFLOPS Scale (10\%);
RS-2025-25442569, AI Star Fellowship Support Program (Sungkyunkwan Univ.) (10\%)),
IITP-ICT Creative Consilience Program grant funded by the Korea government (MSIT) (IITP-2026-RS-2020-II201821) (10\%),
by the National Research Foundation of Korea (NRF) grant funded by the Korea government (MSIT) (RS-2026-25474409) (10\%),
and by Samsung Electronics Co., Ltd. (10\%).

{
\small
\bibliographystyle{unsrt}
\bibliography{main}
}

\newpage
\appendix
\begingroup
\let\citet\cite %
\input{appendix/appendix}
\endgroup

\clearpage
\input{neurips/checklist}

\end{document}

%% file: appendix/appendix.tex
\par\noindent\rule{\textwidth}{2pt}
\begin{table}[!h]
    \vspace{-0.3cm}
    \centering
    \Large
    \begin{tabular}{c}
\textbf{Appendix}
    \end{tabular}
    \vspace{-.5cm}
\end{table}
\par\noindent\rule{\textwidth}{1pt}

\begingroup
\setcounter{tocdepth}{-1}
\tableofcontents
\endgroup
\addtocontents{toc}{\protect\setcounter{tocdepth}{2}}

\newpage

\section{\texorpdfstring{\ours{}}{RecPo}}

\subsection{Procedure Views}
\label{app:view-assumptions}
Procedure views provide a sparse way to govern how procedure records are read. To serve this role, a view should change when the factor it governs changes and remain stable when other factors vary. A view with this property exposes the local effects of skills without enumerating concrete task bindings or full skill sequences. In the main experiments, object- or factor-centric projections provide such views by selecting the observation dimensions of the governed factors. Other functions of the observation can provide such views as well, such as an image embedding for each camera in Appendix~\ref{app:visual-procedure-views} or the fill level of a cup in the real-world experiments of Appendix~\ref{app:realworld}.

Each view enters \ours{} through its interface $(\alpha_v,\delta_v)$, which determines how the procedure records are organized into skill-transition graphs and salience marks. Materialization is defined over this organized structure for an arbitrary view set rather than for a particular choice of views. The interface therefore separates representation from composition and serves as the basis of a modular materialization. Under this modularity, the same materialization rule applies whichever function instantiates the views, and changes to materialization in turn need not redefine how the views are obtained.

\subsection{Procedure Knowledge Base}
\label{app:procedure_kb_details}

{
To compose stored skills through the interfaces of multiple procedure views, the procedure update $U_\Omega$ computes three kinds of evidence from the procedure substrate $(\mathcal{G}_{\Omega,\mathcal{V}},\bar{\mathcal{X}}_{\Omega,\mathcal{V}})$ and the procedure records. The skill-transition graphs provide effect evidence on which view nodes a skill changes. The salience marks provide identification evidence on the view in which the effect of a skill separates it from the other stored skills. The segment endpoints and the skill sequences of the records provide order evidence on where a skill starts and ends and which skill follows it. None of this evidence depends on a task query. Throughout, let $\mathcal{I}_\omega=\{i:\omega_i=\omega\}$ denote the records of skill $\omega$.

\textbf{Effect evidence from the skill-transition graphs.} For each view $v$, $U_\Omega$ canonicalizes the projected starts and endpoints of all records into the nodes $\mathcal{N}_v$. Records of different executions that reach the same view state then share a node, so the effect of a skill becomes a node change that can be counted over its records. $U_\Omega$ computes this partition with DBSCAN and assigns every stored point to a node. The map $c_v$ extends the partition to a new state through the node of the nearest stored point, so task waypoints fall into the same nodes as the records. Over these nodes, $U_\Omega$ groups the edges of $\mathcal{G}_v$ in Eq.~\eqref{eq:view-skill-transition-graph} by skill and start node. Let $\mathcal{I}_{\omega,v}(n)=\{i\in\mathcal{I}_\omega:c_v(z_{i,v})=n\}$ denote the records of $\omega$ that start at node $n\in\mathcal{N}_v$. The empirical effect of $\omega$ at $n$ is
\begin{equation}
\chi_v(n,\omega)=\mathds{1}\!\left[
\frac{\left|\{i\in\mathcal{I}_{\omega,v}(n):c_v(z'_{i,v})\neq c_v(z_{i,v})\}\right|}
{\left|\mathcal{I}_{\omega,v}(n)\right|}\geq\frac{1}{2}
\right],
\label{eq:app-empirical-effect}
\end{equation}
which equals one when at least half of the records of $\omega$ that start at $n$ change the node of view $v$. When no record of $\omega$ starts at $n$, all records of $\omega$ decide the effect. The view-change signature collects the views that $\omega$ changes in at least one record,
\begin{equation}
\Sigma_\omega=\{v\in\mathcal{V}:\exists i\in\mathcal{I}_\omega,\
c_v(z_{i,v})\neq c_v(z'_{i,v})\}.
\label{eq:app-view-change-signature}
\end{equation}

\textbf{Identification evidence from the salience marks.} For each view $v$, the salience mark measures how well the view-local effect of a record predicts its skill label. We instantiate the salience mark $m_{i,v}$ in Eq.~\eqref{eq:correlation-marked-records} as the share of records with skill $\omega_i$ among the nearest neighbors of $e_{i,v}$ in view-local effect, where the neighbors include record $i$ and the share excludes it. Averaging these marks over $\mathcal{I}_\omega$ gives the skill-view predictivity $\lambda_\omega(v)$, and we call $v^*_\omega\in\arg\max_v\lambda_\omega(v)$ the dominant view of $\omega$. Task materialization reads the salience marks only through $\lambda_\omega$.

\textbf{Order evidence from the procedure records.} For order evidence, $U_\Omega$ computes the initiation and termination centroids $\mu^s_\omega$ and $\mu^e_\omega$ from the procedure records without view projection. They are the means of the segment starts $s_i$ and endpoints $s'_i$ over $\mathcal{I}_\omega$. When the records come with their skill sequences, $U_\Omega$ also computes the empirical skill adjacency $\rightarrow_\Omega$ as a connectivity over the stored skills. It contains the ordered pairs $(\omega,\omega')$ of distinct skills whose records are consecutive within one observed execution, independent of the context of that execution.
}

\subsection{Task Materialization}
\label{app:task_materialization_details}

{
Section~\ref{sec:task-query-abstraction} defines task-query abstraction and
segment binding. Using the statistics in
Appendix~\ref{app:procedure_kb_details}, we specify below the support score,
path objective, and algorithmic flow used by these stages.
Table~\ref{tab:app-pkb-statistics} lists where each statistic of Appendix~\ref{app:procedure_kb_details} enters these stages.

\begin{table}[!h]
\centering
\caption{Stored statistics of the Procedure Knowledge Base grouped by evidence and the
materialization steps that read them. The last column names the ablation
axis of Table~\ref{tab:salience-ablation} that changes how a statistic is used. $^{\dagger}$Read only when the records come with their skill sequences.}
\label{tab:app-pkb-statistics}
\footnotesize
\setlength{\tabcolsep}{4pt}
\begin{tabular}{llll}
\toprule
Evidence & Statistic & Used in task materialization & Ablation axis \\
\midrule
\multirow{3}{*}{effect} & $c_v$ & decision points, active views $\mathcal{V}_h$ & -- \\
 & $\chi_v(n,\omega)$ & admissible set $\Omega_h$, decision-point refinement & -- \\
 & $\Sigma_\omega$ & selection of $\Omega_h^0$, decision-point refinement & -- \\
\midrule
identification & $\lambda_\omega$, $v^*_\omega$ & decision points, selection of $\Omega_h^0$, projection $W_{\omega,h}$ & view selection \\
\midrule
\multirow{2}{*}{order} & $\mu^s_\omega$, $\mu^e_\omega$ & candidate support $\rho_h$, path objective $C_h$, decision-point refinement & -- \\
 & $\rightarrow_\Omega$$^{\dagger}$ & final skill order & chaining prior \\
\bottomrule
\end{tabular}
\end{table}
}

{

\textbf{Decision-point refinement.} A node change of Eq.~\eqref{eq:initial-query} yields a decision point only when one changed view maximizes $\lambda_\omega(v)$ for at least one stored skill $\omega$. The first and last steps of $\mathcal{Q}_\tau$ always remain decision points. A skill explains a segment when its signature $\Sigma_\omega$ covers the active views of the segment and its support $\rho$ reaches $\varepsilon_\rho$, where $\rho$ follows Eq.~\eqref{eq:app-aggregate-support} with the active-view indicator of that segment in place of $W_{\omega,h}$. Among these skills, the explaining skill maximizes $\rho/|\Sigma_\omega|$. An interior decision point is removed only when a skill distinct from both adjacent explanations explains the combined segment and its support falls below the weaker adjacent support by at most the tolerance $\varepsilon_{\mathrm{drop}}$. Repeating this removal produces $\bar{\mathcal{Q}}_\tau$, and both settings of each ablation axis share it.

\textbf{Candidate support.}
For a segment $h$ of $\bar{\mathcal{Q}}_\tau$, let $\mathcal{V}_h$ be the active views and
$\mathds{1}_h$ the indicator of their observation dimensions defined in
Section~\ref{sec:task-query-abstraction}. Candidate support is
\begin{equation}
\rho_h(\omega)
=\cos_{W_{\omega,h}}
\left(
q_{j_{h+1}}-q_{j_h},
\mu^e_\omega-\mu^s_\omega
\right),
\label{eq:app-aggregate-support}
\end{equation}
where $\cos_{W_{\omega,h}}$ is the cosine similarity after masking by $W_{\omega,h}$. The mask $W_{\omega,h}$ keeps a subset of the dimensions in $\mathds{1}_h$, and its choice defines the view-selection axis of Table~\ref{tab:salience-ablation}. Candidate-specific selection compares each candidate on its dominant view $v^*_\omega$. When $v^*_\omega$ is inactive in the segment or $\lambda_\omega(v^*_\omega)=0$, the comparison moves to the active views in which the stored records show one irreversible change, and to all of $\mathds{1}_h$ when no such view is active. Uniform selection sets $W_{\omega,h}=\mathds{1}_h$ for every candidate.

The admissible set $\Omega_h$ of segment $h$ contains the skills whose empirical effect in Eq.~\eqref{eq:app-empirical-effect} changes the start node of at least one active view. When the signature $\Sigma_\omega$ of some candidate covers $\mathcal{V}_h$, both settings consider only the candidates with such a signature. Otherwise, candidate-specific selection keeps only the candidates whose dominant view is active with $\lambda_\omega(v^*_\omega)>0$, and uniform selection keeps all of them. The selected candidates form $\Omega_h^0$ and satisfy $\rho_h(\omega)\geq\varepsilon_\rho$ for the support threshold $\varepsilon_\rho$.

\textbf{Path objective.} The path operator in Eq.~\eqref{eq:segment-binding} orders the skills of $\Omega_h^0$ by approximately minimizing
\begin{equation}
\begin{aligned}
C_h(\omega_{1:L_h})={}&d_h(q_{j_h},\mu^s_{\omega_1})
+\sum_{l=1}^{L_h-1}d_h(\mu^e_{\omega_l},\mu^s_{\omega_{l+1}})
+d_h(\mu^e_{\omega_{L_h}},q_{j_{h+1}}),
\end{aligned}
\label{eq:app-path-objective}
\end{equation}
where $d_h$ is the $L_2$ distance on the dimensions of $\mathds{1}_h$. A bounded bidirectional beam search expands partial sequences from both endpoints and orders every skill in $\Omega_h^0$, so $L_h=|\Omega_h^0|$.

\textbf{Composition and chaining prior.} Task-level composition emits each selected skill once at its earliest segment binding and returns one executable skill chain. When the records come with their skill sequences, the empirical skill adjacency $\rightarrow_\Omega$ reorders this chain only when it admits exactly one order that covers all chained skills. The reorder applies to the whole chain and can cross segment boundaries. It defines the chaining-prior axis of Table~\ref{tab:salience-ablation}, and chaining without the prior keeps the composed order.

\begin{algorithm}[!h]
\caption{Task materialization in \ours{}}
\label{alg:app-task-materialization}
\begin{algorithmic}
\footnotesize
\Require task query $\mathcal{Q}_\tau$ and Procedure Knowledge Base
$\mathcal{K}_\Omega$
\Ensure materialized task $\widehat{\mathcal{Q}}_\tau$

\BeginBox[fill=mygreen!8]
\LComment{\color{mygreen} Abstract the task query}
\State Project $\mathcal{Q}_\tau$ through the supplied views and collect
node changes containing a view that maximizes $\lambda_\omega(v)$ for a stored
skill.
\State Refine the decision points with stored segment support to form
$\bar{\mathcal{Q}}_\tau$.
\EndBox

\vspace{5pt}

\BeginBox[fill=white]
\LComment{\color{mygreen} Bind the abstracted segments}
\For{each segment $(q_{j_h},q_{j_{h+1}})$ in
$\bar{\mathcal{Q}}_\tau$}
    \State Identify the active views $\mathcal{V}_h$ and indicator
    $\mathds{1}_h$.
    \State Collect the skills with $\chi_v=1$ at the active-view start nodes.
    \State Select $\Omega_h^0$ from {the admissible set
    $\Omega_h$} satisfying $\rho_h(\omega)\geq\varepsilon_\rho$.
    \State Bind the segment to $\hat{\omega}_{h,1:L_h}$ by approximately
    minimizing $C_h$ in Eq.~\eqref{eq:app-path-objective} over $\Omega_h^0$.
\EndFor
\EndBox

\BeginBox[fill=white]
\LComment{\color{mygreen} Compose the executable task}
\State Emit every selected skill once at its earliest segment binding.
\State Reorder these bindings only if $\rightarrow_\Omega$ defines a unique
covering order.
\EndBox

\State \Return $\widehat{\mathcal{Q}}_\tau$ as defined in
Eq.~\eqref{eq:materialized-task}.
\end{algorithmic}
\end{algorithm}

The simulation protocol executes the fixed sequence returned by
Algorithm~\ref{alg:app-task-materialization}. This design isolates task
abstraction and composition from feedback control. The reported results
therefore characterize materialization before execution-layer disturbance
detection or recovery is introduced.
Table~\ref{tab:app-hyperparameters} lists the hyperparameters of the Procedure Knowledge Base and task materialization.

\begin{table}[!h]
\centering
\caption{Hyperparameters of the Procedure Knowledge Base and task materialization.}
\label{tab:app-hyperparameters}
\footnotesize
\begin{tabular}{llc}
\toprule
Stage & Hyperparameter & Value \\
\midrule
\multirow{3}{*}{Procedure Knowledge Base} & DBSCAN neighborhood radius & $0.10$ \\
 & DBSCAN minimum sample count & $20$ \\
 & salience nearest neighbors & $32$ \\
\midrule
\multirow{3}{*}{Task materialization} & support threshold $\varepsilon_\rho$ & $0.10$ \\
 & refinement tolerance $\varepsilon_{\mathrm{drop}}$ & $0.05$ \\
 & beam width & $8$ \\
\bottomrule
\end{tabular}
\end{table}

}

\section{Experiment Details}
\label{app:experiment-details}
For the camera-ready version, we update the results by extending the evaluation scope of the experiments that were limited in the submitted version and the rebuttal, and by re-measuring task-inference times with a codebase that uses optimized clustering and nearest-neighbor retrieval algorithms.

\subsection{Environment}
\textbf{Procedure records.}
We populate the procedure knowledge base with the length-$4$ Franka Kitchen realizations provided by \citet{gupta2019relay, fu2020d4rl}.
Each realization starts from the common initial configuration and executes four of seven subtasks in the relative order of the target task in Table~\ref{tab:app-target-order}.
Because the realizations instantiate different subtask subsets, the same skill may be executed after different preceding subtasks have been completed, yielding distinct full-state contexts for the same view-local effect.
We refer to this variation in the contexts supporting reusable skill effects as \emph{skill-context diversity}.

\begin{figure}[!ht]
    \centering
    \includegraphics[width=\linewidth]{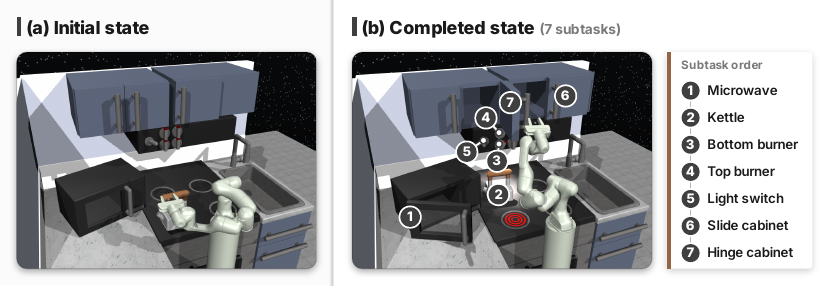}
    \caption{Franka Kitchen environment. (a) The common initial configuration from which every realization starts. (b) The state after the held-out length-$7$ target completes all seven subtasks, numbered in the target order.}
    \label{fig:kitchen-environment}
\end{figure}

\textbf{Held-out length-$7$.}
The held-out length-$7$ target completes all seven subtasks, while its complete realization is excluded from the procedure knowledge base.
Figure~\ref{fig:kitchen-environment} shows the common initial configuration and the state after all seven subtasks are completed.
It therefore requires skill-induced effects observed in the length-$4$ records to be reused across joint contexts absent from any stored realization.
This shared relative order constrains the subtask order, leaving task materialization to identify transition evidence for the demanded effects and compose the corresponding skills.
We further increase the \textit{waypoint gap} $H$ to remove intermediate task-state observations without changing this ordering structure.
As $H$ grows, a single observed waypoint transition may span multiple subtask effects, reducing the task-side evidence available for binding them to stored procedure records.
The held-out length-$7$ setting thus evaluates composition of reusable effects across unseen context combinations under increasingly sparse task evidence.

{
\textbf{KB construction details.}
A procedure realization is a sequence of $30$-dimensional joint configurations.
Each configuration holds the $9$ robot joints ($7$ arm joints and $2$ gripper fingers) and the $21$ object joints.
Each subtask has a goal configuration of its object joints~\cite{fu2020d4rl}, and a subtask is complete when its object joints are within the Euclidean distance $0.3$ of this goal.
Each observation has the label of the subtask that the robot executes.
The label changes after the subtask is complete and its object stops moving.
We split each realization into maximal runs of the same label.
Each run gives one procedure record $x_i=(s_i,s'_i,\omega_i,k_i)$ in Eq.~\eqref{eq:procedure-record}.
Thus a record starts where the previous record ends, and its end observation shows the settled effect of its skill.
The motion of the robot toward the next object belongs to the next record.
We use the labels only for this split.
The knowledge base does not receive a task identity or a target order.

We group the realizations by their length-$4$ subtask order (Table~\ref{tab:app-procedure-orders}).
The tables name each subtask by its object as \texttt{MW} for the microwave, \texttt{KT} for the kettle, \texttt{BB} for the bottom burner, \texttt{TB} for the top burner, \texttt{LS} for the light switch, \texttt{SC} for the slide cabinet, and \texttt{HC} for the hinge cabinet.
$24$ orders come from the original Franka Kitchen demonstrations~\cite{gupta2019relay}, and we add $8$ orders.
We build the realizations of the added orders by stitching subtask segments of the original demonstrations.
Each order gives $18$ to $25$ realizations.
A skill-context diversity setting $K$ contains the orders with the IDs $1$ to $K$.
Thus a larger $K$ only adds records to a smaller $K$.
\begin{itemize}[leftmargin=*,itemsep=0pt,topsep=2pt]
\item $K=4$: $93$ realizations, $372$ procedure records.
\item $K=8$: $193$ realizations, $772$ procedure records.
\item $K=16$: $393$ realizations, $1{,}572$ procedure records.
\item $K=32$: $776$ realizations, $3{,}104$ procedure records.
\end{itemize}
We build the knowledge base one time for each pair of $K$ and view configuration.
All waypoint gaps, seeds, and evaluation episodes use the same knowledge base.

\begin{table}[!ht]
\centering
\small
\caption{Length-$4$ subtask orders of the procedure realizations. A number gives the step of the subtask in the order. Unshaded rows are the $24$ orders of the original Franka Kitchen demonstrations~\cite{gupta2019relay}, and shaded rows are the $8$ added orders.}
\label{tab:app-procedure-orders}
\setlength{\tabcolsep}{4pt}
\renewcommand{\arraystretch}{1.05}
\begin{tabular}{r*{7}{c}@{\hspace{18pt}}r*{7}{c}}
\toprule
ID & \texttt{MW} & \texttt{KT} & \texttt{BB} & \texttt{TB} & \texttt{LS} & \texttt{SC} & \texttt{HC} & ID & \texttt{MW} & \texttt{KT} & \texttt{BB} & \texttt{TB} & \texttt{LS} & \texttt{SC} & \texttt{HC} \\
\midrule
\cellcolor{gray!15}$1$ & \cellcolor{gray!15}$\cdot$ & \cellcolor{gray!15}$\cdot$ & \cellcolor{gray!15}$1$ & \cellcolor{gray!15}$\cdot$ & \cellcolor{gray!15}$2$ & \cellcolor{gray!15}$3$ & \cellcolor{gray!15}$4$ & \cellcolor{gray!15}$17$ & \cellcolor{gray!15}$\cdot$ & \cellcolor{gray!15}$1$ & \cellcolor{gray!15}$\cdot$ & \cellcolor{gray!15}$2$ & \cellcolor{gray!15}$3$ & \cellcolor{gray!15}$\cdot$ & \cellcolor{gray!15}$4$ \\
$2$ & $1$ & $2$ & $\cdot$ & $3$ & $\cdot$ & $\cdot$ & $4$ & $18$ & $1$ & $\cdot$ & $2$ & $\cdot$ & $3$ & $4$ & $\cdot$ \\
$3$ & $\cdot$ & $\cdot$ & $1$ & $2$ & $3$ & $4$ & $\cdot$ & \cellcolor{gray!15}$19$ & \cellcolor{gray!15}$\cdot$ & \cellcolor{gray!15}$1$ & \cellcolor{gray!15}$\cdot$ & \cellcolor{gray!15}$2$ & \cellcolor{gray!15}$\cdot$ & \cellcolor{gray!15}$3$ & \cellcolor{gray!15}$4$ \\
$4$ & $1$ & $2$ & $3$ & $\cdot$ & $\cdot$ & $\cdot$ & $4$ & $20$ & $1$ & $\cdot$ & $2$ & $\cdot$ & $\cdot$ & $3$ & $4$ \\
$5$ & $\cdot$ & $1$ & $\cdot$ & $2$ & $3$ & $4$ & $\cdot$ & $21$ & $\cdot$ & $1$ & $2$ & $\cdot$ & $3$ & $4$ & $\cdot$ \\
\cellcolor{gray!15}$6$ & \cellcolor{gray!15}$1$ & \cellcolor{gray!15}$\cdot$ & \cellcolor{gray!15}$2$ & \cellcolor{gray!15}$\cdot$ & \cellcolor{gray!15}$3$ & \cellcolor{gray!15}$\cdot$ & \cellcolor{gray!15}$4$ & $22$ & $1$ & $\cdot$ & $\cdot$ & $2$ & $3$ & $\cdot$ & $4$ \\
\cellcolor{gray!15}$7$ & \cellcolor{gray!15}$1$ & \cellcolor{gray!15}$2$ & \cellcolor{gray!15}$\cdot$ & \cellcolor{gray!15}$3$ & \cellcolor{gray!15}$\cdot$ & \cellcolor{gray!15}$4$ & \cellcolor{gray!15}$\cdot$ & $23$ & $\cdot$ & $1$ & $2$ & $\cdot$ & $\cdot$ & $3$ & $4$ \\
\cellcolor{gray!15}$8$ & \cellcolor{gray!15}$\cdot$ & \cellcolor{gray!15}$\cdot$ & \cellcolor{gray!15}$1$ & \cellcolor{gray!15}$2$ & \cellcolor{gray!15}$3$ & \cellcolor{gray!15}$\cdot$ & \cellcolor{gray!15}$4$ & $24$ & $1$ & $\cdot$ & $2$ & $3$ & $4$ & $\cdot$ & $\cdot$ \\
$9$ & $1$ & $2$ & $3$ & $\cdot$ & $\cdot$ & $4$ & $\cdot$ & $25$ & $1$ & $2$ & $\cdot$ & $\cdot$ & $3$ & $\cdot$ & $4$ \\
$10$ & $\cdot$ & $1$ & $\cdot$ & $\cdot$ & $2$ & $3$ & $4$ & $26$ & $\cdot$ & $1$ & $2$ & $3$ & $\cdot$ & $\cdot$ & $4$ \\
$11$ & $1$ & $\cdot$ & $2$ & $3$ & $\cdot$ & $\cdot$ & $4$ & \cellcolor{gray!15}$27$ & \cellcolor{gray!15}$1$ & \cellcolor{gray!15}$\cdot$ & \cellcolor{gray!15}$\cdot$ & \cellcolor{gray!15}$2$ & \cellcolor{gray!15}$3$ & \cellcolor{gray!15}$4$ & \cellcolor{gray!15}$\cdot$ \\
$12$ & $1$ & $2$ & $\cdot$ & $\cdot$ & $3$ & $4$ & $\cdot$ & $28$ & $\cdot$ & $1$ & $2$ & $3$ & $4$ & $\cdot$ & $\cdot$ \\
$13$ & $\cdot$ & $\cdot$ & $1$ & $2$ & $\cdot$ & $3$ & $4$ & $29$ & $1$ & $2$ & $\cdot$ & $\cdot$ & $\cdot$ & $3$ & $4$ \\
$14$ & $1$ & $2$ & $\cdot$ & $3$ & $4$ & $\cdot$ & $\cdot$ & \cellcolor{gray!15}$30$ & \cellcolor{gray!15}$1$ & \cellcolor{gray!15}$\cdot$ & \cellcolor{gray!15}$\cdot$ & \cellcolor{gray!15}$2$ & \cellcolor{gray!15}$\cdot$ & \cellcolor{gray!15}$3$ & \cellcolor{gray!15}$4$ \\
$15$ & $\cdot$ & $1$ & $2$ & $\cdot$ & $3$ & $\cdot$ & $4$ & $31$ & $\cdot$ & $1$ & $2$ & $3$ & $\cdot$ & $4$ & $\cdot$ \\
$16$ & $1$ & $\cdot$ & $2$ & $3$ & $\cdot$ & $4$ & $\cdot$ & $32$ & $1$ & $\cdot$ & $\cdot$ & $\cdot$ & $2$ & $3$ & $4$ \\
\bottomrule
\end{tabular}
\end{table}

The procedure views follow the object factors that D4RL defines for the Franka Kitchen state~\cite{fu2020d4rl}.
Each factor is a fixed set of state dimensions, and Table~\ref{tab:app-view-configs} gives these dimensions.
These factors are not privileged information.
Counterfactual tests on interaction data can identify the state factors that an action or an entity changes~\cite{pitis2020counterfactual,caiac2024,chuck2025null}.
Every configuration keeps the robot view and the kettle view, and it merges the other six object factors into $V-2$ views.

\begin{table}[!ht]
\centering
\small
\caption{State dimensions of the view factors (top) and views of the seven view configurations (bottom). \texttt{RB} is the robot. The robot view also holds the dimensions $9$--$10$ of a burner that no subtask uses. The dimensions $13$--$14$ of the other unused burner are in no view.}
\label{tab:app-view-configs}
\renewcommand{\arraystretch}{1.1}
\begin{tabular}{lcccccccc}
\toprule
Factor & \texttt{RB} & \texttt{BB} & \texttt{TB} & \texttt{LS} & \texttt{SC} & \texttt{HC} & \texttt{MW} & \texttt{KT} \\
\midrule
Dimensions & $0$--$10$ & $11$--$12$ & $15$--$16$ & $17$--$18$ & $19$ & $20$--$21$ & $22$ & $23$--$29$ \\
\bottomrule
\end{tabular}

\vspace{6pt}
\begin{tabular}{cll}
\toprule
$V$ & Fixed views & Merged object views \\
\midrule
$8$ & $\{$\texttt{RB}$\}$\enspace $\{$\texttt{KT}$\}$ & $\{$\texttt{MW}$\}$\enspace $\{$\texttt{BB}$\}$\enspace $\{$\texttt{TB}$\}$\enspace $\{$\texttt{LS}$\}$\enspace $\{$\texttt{SC}$\}$\enspace $\{$\texttt{HC}$\}$ \\
$5$ & $\{$\texttt{RB}$\}$\enspace $\{$\texttt{KT}$\}$ & $\{$\texttt{LS}, \texttt{SC}$\}$\enspace $\{$\texttt{MW}, \texttt{HC}$\}$\enspace $\{$\texttt{BB}, \texttt{TB}$\}$ \\
$5$ & $\{$\texttt{RB}$\}$\enspace $\{$\texttt{KT}$\}$ & $\{$\texttt{BB}, \texttt{HC}$\}$\enspace $\{$\texttt{TB}, \texttt{LS}$\}$\enspace $\{$\texttt{MW}, \texttt{SC}$\}$ \\
$5$ & $\{$\texttt{RB}$\}$\enspace $\{$\texttt{KT}$\}$ & $\{$\texttt{MW}, \texttt{SC}$\}$\enspace $\{$\texttt{LS}, \texttt{HC}$\}$\enspace $\{$\texttt{BB}, \texttt{TB}$\}$ \\
$4$ & $\{$\texttt{RB}$\}$\enspace $\{$\texttt{KT}$\}$ & $\{$\texttt{MW}, \texttt{LS}, \texttt{SC}, \texttt{HC}$\}$\enspace $\{$\texttt{BB}, \texttt{TB}$\}$ \\
$4$ & $\{$\texttt{RB}$\}$\enspace $\{$\texttt{KT}$\}$ & $\{$\texttt{BB}, \texttt{TB}, \texttt{LS}, \texttt{HC}$\}$\enspace $\{$\texttt{MW}, \texttt{SC}$\}$ \\
$3$ & $\{$\texttt{RB}$\}$\enspace $\{$\texttt{KT}$\}$ & $\{$\texttt{MW}, \texttt{BB}, \texttt{TB}, \texttt{LS}, \texttt{SC}, \texttt{HC}$\}$ \\
\bottomrule
\end{tabular}
\end{table}

\textbf{Task specification.}
The task demonstration is one held-out length-$7$ realization with $383$ observations.
Its subtask order is the target order in Table~\ref{tab:app-target-order}.
We split it with the same rule as the procedure realizations.
The initial observation and the end observations of the seven runs give eight boundary observations with the indices $\{0,\ldots,7\}$.
For a waypoint gap $H$, the task record keeps the boundary indices $\{0,H,2H,\ldots\}$ that are not larger than $7$.
If the last kept index is not $7$, the task record also keeps the final observation.
At $H=7$, only the initial and the final observations remain.
All methods receive the same waypoint array, and its first row is the initial observation of the episode.
No method receives subtask labels or skill identities from the task demonstration.

\begin{table}[!ht]
\centering
\small
\caption{Subtask order of the held-out length-$7$ target task.}
\label{tab:app-target-order}
\setlength{\tabcolsep}{5pt}
\begin{tabular}{cccccccc}
\toprule
Step & $1$ & $2$ & $3$ & $4$ & $5$ & $6$ & $7$ \\
\midrule
Subtask & \texttt{MW} & \texttt{KT} & \texttt{BB} & \texttt{TB} & \texttt{LS} & \texttt{SC} & \texttt{HC} \\
\bottomrule
\end{tabular}
\end{table}

\textbf{Executable skill construction.}
The evaluation executes a materialized skill sequence with a library of executable skills, one skill for each kitchen object.
A skill is one kinematic transition.
It sets the joints of its object to the goal configuration of its subtask and sets the $9$ robot joints to a robot pose sampled from the end observations of the records of the same skill in the length-$4$ realizations.
All other joints keep their values, so the outcome of an episode depends only on the selected skills and their order.
After each skill, the environment returns the joint configuration with the sensor noise of Franka Kitchen.
The evaluation thus measures procedure reconfiguration.
}

\subsection{Baseline Implementations} %
\label{app:baseline-scope}
All baselines use the same stored skill records, supplied procedure views,
and executable skill library.

\textbf{\texttt{Skill Retrieval} (\texttt{SR}).}
\texttt{SR} performs transition-level retrieval and reranking.
For each task-waypoint transition, it scores each skill by matching the requested transition to the closest stored record of that skill under each supplied procedure view.
The view-wise scores are averaged to obtain a single retrieval score for each skill, and the highest-scoring skill is selected independently for each waypoint transition.

\textbf{\texttt{Skill Sequence Retrieval} (\texttt{SSR}).}
\texttt{SSR} extends the retrieval unit from individual skill records to stored skill subsequences.
For each length-$4$ procedure realization, we construct retrieval items from its contiguous skill subsequences of lengths $1$, $2$, and $3$, preserving their observed execution order.
Each subsequence is represented under the same supplied procedure views as \texttt{SR} and matched to a task-waypoint segment using the same multi-view scoring criterion.
For each task segment, \texttt{SSR} retrieves the highest-scoring stored subsequence.
This baseline evaluates whether retrieval over previously observed short skill subsequences is sufficient to recover the target behavior from sparse task evidence.

\textbf{\texttt{Skill Chaining} (\texttt{SC}).}
\texttt{SC} instantiates the structured procedural-representation family of skill chaining and skill-graph methods~\cite{konidaris2009skill, bagaria2021skill} within the shared reconfiguration interface above.
It uses the supplied procedure views when matching task waypoints to stored record endpoints, giving it access to the same factorized procedure information as \ours{}.

The distinction lies in how this evidence is organized for composition.
\texttt{SC} summarizes the records associated with each skill into canonical initiation and termination states in the procedure-view space, thereby representing procedural connectivity at the skill level.
Given a task query, it scores candidate skills by agreement with the inferred waypoint changes and constructs a skill-transition graph whose edges are weighted by termination-to-initiation distance.
A bounded path search over this graph returns the executable skill sequence.
\texttt{SC} therefore tests whether the supplied factorized procedure information remains sufficient after record-level evidence is consolidated into skill-level connectivity.
In contrast, \ours{} retains record-derived view-local transition evidence and resolves the task-relevant composition during materialization.

\subsection{Computing Details}
Table~\ref{tab:computing-environment} lists the compute resources and core library versions.
\ours{} uses NumPy for numerical array operations and RAPIDS-based GPU libraries for scalable clustering and nearest-neighbor retrieval.
We use scikit-learn and UMAP for CPU-based clustering and dimensionality reduction.

\begin{table}[!h]
\centering
\footnotesize
\renewcommand{\arraystretch}{1.15}
\setlength{\tabcolsep}{4pt}
\caption{Compute resources and core library versions.}
\vskip 0.1in
\label{tab:computing-environment}
\begin{tabular*}{\textwidth}{@{\extracolsep{\fill}\hspace{\tabcolsep}}ll|llll@{\hspace{\tabcolsep}}}
\toprule
\multicolumn{2}{l|}{\textbf{Compute resources}} & \multicolumn{4}{l}{\textbf{Clustering and numerical libraries}} \\
\midrule
CPU & AMD Ryzen Threadripper PRO 9975WX (32-core) & \texttt{numpy} & 2.2.6 & \texttt{cuml} & 25.12.0 \\
GPU & 2$\times$ NVIDIA GeForce RTX 5090 & \texttt{scipy} & 1.16.3 & \texttt{cuvs} & 25.12.0 \\
OS & Ubuntu 24.04.3 LTS & \texttt{scikit-learn} & 1.7.2 & \texttt{cugraph} & 25.12.2 \\
CUDA & 12.8 & \texttt{umap-learn} & 0.5.11 & \texttt{cupy} & 14.0.1 \\
Driver & 570.211.01 & \texttt{pynndescent} & 0.6.0 & & \\
\bottomrule
\end{tabular*}
\end{table}

\subsection{Evaluation Protocol and Aggregation}
\label{app:evaluation-protocol}

{
The procedure-support and task-record-sparsity analyses use the sweep in Table~\ref{tab:shared-evaluation-conditions}.
The stored length-$4$ realizations provide the skill-context diversity settings.
The seven procedure-view configurations use distinct factor-merging assignments under the same view-construction rule.
The $V=3,4,5,8$ levels contain $1,2,3,1$ configurations, respectively.
Both analyses report marginals of this sweep.
The task-input conditions are described in Section~\ref{app:baseline-scope}.

Experiment condition tables use the notation in Table~\ref{tab:supplementary-notation}.
Counts cover the listed configurations for each method and experimental variant.
For randomized view construction, we first average within each construction seed.
We then report the mean and sample standard deviation across construction seeds.
}

\begin{table}[!h]
\centering
\small
\caption{Notation for experiment condition tables.}
\label{tab:supplementary-notation}
\renewcommand{\arraystretch}{1.15}
\begin{tabular*}{\textwidth}{@{\extracolsep{\fill}\hspace{\tabcolsep}}lp{0.76\textwidth}@{\hspace{\tabcolsep}}}
\toprule
Notation & Meaning \\
\midrule
$K$, $H$, $V$ & Skill-context diversity, waypoint gap, and view resolution. \\
$S_e$ & Number of evaluation seeds, each with $100$ episodes. \\
$N_{\mathrm{cfg}}$ & Number of listed configurations. \\
$N_{\mathrm{ep}}$ & Total quality-evaluation episodes, $100\,N_{\mathrm{cfg}}S_e$, with $\mathrm{k}=10^3$. \\
$\dagger$ & An axis averaged over its listed configurations. Additional averaging for Mean is stated in each caption. \\
-- & Not applicable. \\
\bottomrule
\end{tabular*}
\end{table}

\begin{table}[!h]
\centering
\small
\caption{196-configuration sweep for procedure-support and task-record-sparsity evaluation. $N_{\mathrm{ep}}$ counts quality-evaluation episodes per method.}
\label{tab:shared-evaluation-conditions}
\renewcommand{\arraystretch}{1.15}
\setlength{\tabcolsep}{3pt}
\begin{tabular*}{\textwidth}{@{\extracolsep{\fill}\hspace{\tabcolsep}}ccccc@{\hspace{\tabcolsep}}}
\toprule
$K$ & $H$ & Views & $N_{\mathrm{cfg}}$ & $N_{\mathrm{ep}}$ \\
\midrule
$\{4,8,16,32\}$ & $\{1,\ldots,7\}$ & \makecell[c]{$V\in\{3,4,5,8\}$\\$7$ configurations total} & $4\times7\times7$ & $196\mathrm{k}$ \\
\bottomrule
\end{tabular*}
\end{table}

\textbf{Ordered-completion evaluation.}
{An episode ends after at most $12$ skill executions, and ordered completion (OC) counts the subtasks completed in the target order.}
{Each configuration uses $S_e=10$ evaluation seeds and $100$ episodes per seed, yielding $196{,}000$ episodes per method.}
We first average all episodes across the ten seeds within each configuration and then weight configurations equally.
Marginal results fix one axis and average over the remaining configurations.
Accordingly, each $K$ value averages $49$ configurations, each $H$ value averages $28$, and the four $V$ levels average $28$, $56$, $84$, and $28$ configurations, respectively.

\textbf{Timing evaluation.}
Timing is measured in a separate run over the same configurations.
Each configuration uses two seeds and $100$ repetitions per seed.
We exclude the first repetition of each seed as warmup, leaving $198$ measurements per configuration and $38{,}808$ measurements per method.
Materialization is recomputed for every repetition, while the procedure knowledge base is reused.
The reported task-inference time includes task-query abstraction, materialization, and all policy-prediction calls within an episode, while procedure-knowledge-base construction, file loading, environment communication, and physical execution are excluded.
Quality and timing use the same configuration weighting while being measured independently.

\textbf{Aggregation and variability.}
For the sweep above, error terms for both \textit{ordered completion} and timing are sample standard deviations across configuration means.
Marginals hold the displayed axis fixed and vary the remaining configuration dimensions, including distinct view configurations at the same $V$.
These deviations describe configuration heterogeneity and are not confidence intervals.

\subsection{Main Evaluation Results}
\label{app:main-evaluation-results}

Table~\ref{tab:rq1_headline_average} summarizes \textit{ordered completion} and
materialization and prediction time over the sweep in
Table~\ref{tab:shared-evaluation-conditions}. The following subsections detail
\textit{procedure support} and \textit{task-record sparsity} separately.

\begin{table}[!h]
\centering
\caption{Mean \textit{ordered completion} and materialization and prediction time over the procedure-support and task-record-sparsity sweep. Error terms describe variation across configuration means under Section~\ref{app:evaluation-protocol}.}
\vskip 0.1in
\label{tab:rq1_headline_average}
\label{tab:rq2_headline_average}
\renewcommand{\arraystretch}{1.15}
\setlength{\tabcolsep}{4pt}
\begingroup
\footnotesize
\renewcommand{\arraystretch}{1.15}
\setlength{\tabcolsep}{3pt}
\begin{tabular*}{\textwidth}{@{\extracolsep{\fill}\hspace{\tabcolsep}}ccccc@{\hspace{\tabcolsep}}}
\toprule
\multicolumn{3}{l}{\textbf{Experiment Config}} & \multicolumn{2}{l}{} \\
\midrule
$K^{\dagger}$ & $H^{\dagger}$ & Views$^{\dagger}$ & $N_{\mathrm{cfg}}$ & $N_{\mathrm{ep}}$ \\
\midrule
$\{4,8,16,32\}$ & $\{1,\ldots,7\}$ & \makecell[c]{$V\in\{3,4,5,8\}$\\$7$ configurations total} & $4\times7\times7$ & $196\mathrm{k}$ \\
\bottomrule
\end{tabular*}
\endgroup
\par\smallskip
\begin{adjustbox}{width=\textwidth}
\begin{tabular}{c|cc|cc|cc|>{\columncolor{gray!25}}c>{\columncolor{gray!25}}c}
\toprule
\multicolumn{3}{l}{\textbf{Results}} & \multicolumn{6}{l}{} \\
\midrule
& \multicolumn{2}{c|}{\makecell[c]{\texttt{Skill Retrieval}\\(\texttt{SR})}}
& \multicolumn{2}{c|}{\makecell[c]{\texttt{Skill Sequence}\\\texttt{Retrieval} (\texttt{SSR})}}
& \multicolumn{2}{c|}{\makecell[c]{\texttt{Skill Chaining}\\(\texttt{SC})}}
& \multicolumn{2}{c}{\makecell[c]{\ours{}\\(ours)}} \\
\cmidrule(lr){2-3}\cmidrule(lr){4-5}\cmidrule(lr){6-7}\cmidrule(lr){8-9}
Average
& \textsc{OC}$\uparrow${\scriptsize(\%)} & Time$\downarrow${\scriptsize(ms)}
& \textsc{OC}$\uparrow${\scriptsize(\%)} & Time$\downarrow${\scriptsize(ms)}
& \textsc{OC}$\uparrow${\scriptsize(\%)} & Time$\downarrow${\scriptsize(ms)}
& {\cellcolor{white!25}\textsc{OC}$\uparrow${\scriptsize(\%)}} & {\cellcolor{white!25}Time$\downarrow${\scriptsize(ms)}} \\
\midrule
Overall & 8.1{\color{gray}\scriptsize$\pm$15.9} & 2.4{\color{gray}\scriptsize$\pm$1.0} & 24.7{\color{gray}\scriptsize$\pm$29.5} & 6.7{\color{gray}\scriptsize$\pm$2.0} & 51.8{\color{gray}\scriptsize$\pm$34.3} & 0.9{\color{gray}\scriptsize$\pm$0.7} & 91.9{\color{gray}\scriptsize$\pm$15.9} & 1.0{\color{gray}\scriptsize$\pm$0.7} \\
\bottomrule
\end{tabular}
\end{adjustbox}
\end{table}

\subsubsection{Procedure-Support Evaluation}
\label{app:procedure-support}

We examine \textit{procedure support} by varying \textit{skill-context diversity} and \textit{view
resolution}. The marginal results below show how completion and computation vary with
\textit{procedure support}.

\textbf{Skill-context diversity.}
Table~\ref{tab:rq1_skill_context_diversity} holds $K$ fixed while averaging
all supplied view configurations and \textit{waypoint gaps}. It therefore measures
how performance changes as more length-$4$ realizations provide \textit{procedure
support} for the same held-out target.

\begin{table}[!h]
\centering
\caption{\textit{Ordered completion} and materialization and prediction time by \textit{skill-context diversity} $K$. Both error terms describe variation across configuration means.}
\vskip 0.1in
\label{tab:rq1_skill_context_diversity}
\renewcommand{\arraystretch}{1.15}
\setlength{\tabcolsep}{4pt}
\begingroup
\footnotesize
\renewcommand{\arraystretch}{1.15}
\setlength{\tabcolsep}{3pt}
\begin{tabular*}{\textwidth}{@{\extracolsep{\fill}\hspace{\tabcolsep}}ccccc@{\hspace{\tabcolsep}}}
\toprule
\multicolumn{3}{l}{\textbf{Experiment Config}} & \multicolumn{2}{l}{} \\
\midrule
$K$ & $H^{\dagger}$ & Views$^{\dagger}$ & $N_{\mathrm{cfg}}$ & $N_{\mathrm{ep}}$ \\
\midrule
$\{4,8,16,32\}$ & $\{1,\ldots,7\}$ & \makecell[c]{$V\in\{3,4,5,8\}$\\$7$ configurations total} & $4\times7\times7$ & $196\mathrm{k}$ \\
\bottomrule
\end{tabular*}
\endgroup
\par\smallskip
\begin{adjustbox}{width=\textwidth}
\begin{tabular}{c|cc|cc|cc|>{\columncolor{gray!25}}c>{\columncolor{gray!25}}c}
\toprule
\multicolumn{3}{l}{\textbf{Results}} & \multicolumn{6}{l}{} \\
\midrule
& \multicolumn{2}{c|}{\makecell[c]{\texttt{Skill Retrieval}\\(\texttt{SR})}}
& \multicolumn{2}{c|}{\makecell[c]{\texttt{Skill Sequence}\\\texttt{Retrieval} (\texttt{SSR})}}
& \multicolumn{2}{c|}{\makecell[c]{\texttt{Skill Chaining}\\(\texttt{SC})}}
& \multicolumn{2}{c}{\makecell[c]{\ours{}\\(ours)}} \\
\cmidrule(lr){2-3}\cmidrule(lr){4-5}\cmidrule(lr){6-7}\cmidrule(lr){8-9}
$K$
& \textsc{OC}$\uparrow${\scriptsize(\%)} & Time$\downarrow${\scriptsize(ms)}
& \textsc{OC}$\uparrow${\scriptsize(\%)} & Time$\downarrow${\scriptsize(ms)}
& \textsc{OC}$\uparrow${\scriptsize(\%)} & Time$\downarrow${\scriptsize(ms)}
& {\cellcolor{white!25}\textsc{OC}$\uparrow${\scriptsize(\%)}} & {\cellcolor{white!25}Time$\downarrow${\scriptsize(ms)}} \\
\midrule
4 & 9.6{\color{gray}\scriptsize$\pm$15.9} & 1.6{\color{gray}\scriptsize$\pm$0.5} & 20.5{\color{gray}\scriptsize$\pm$23.8} & 5.3{\color{gray}\scriptsize$\pm$1.6} & 56.9{\color{gray}\scriptsize$\pm$32.0} & 0.5{\color{gray}\scriptsize$\pm$0.2} & 81.0{\color{gray}\scriptsize$\pm$21.1} & 0.5{\color{gray}\scriptsize$\pm$0.2} \\
8 & 7.1{\color{gray}\scriptsize$\pm$14.4} & 1.9{\color{gray}\scriptsize$\pm$0.6} & 18.0{\color{gray}\scriptsize$\pm$23.0} & 6.5{\color{gray}\scriptsize$\pm$1.8} & 50.7{\color{gray}\scriptsize$\pm$36.0} & 0.7{\color{gray}\scriptsize$\pm$0.2} & 92.7{\color{gray}\scriptsize$\pm$14.6} & 0.7{\color{gray}\scriptsize$\pm$0.2} \\
16 & 7.1{\color{gray}\scriptsize$\pm$14.6} & 2.5{\color{gray}\scriptsize$\pm$0.6} & 21.6{\color{gray}\scriptsize$\pm$33.7} & 6.9{\color{gray}\scriptsize$\pm$1.9} & 52.2{\color{gray}\scriptsize$\pm$33.6} & 0.8{\color{gray}\scriptsize$\pm$0.3} & 96.2{\color{gray}\scriptsize$\pm$12.3} & 0.9{\color{gray}\scriptsize$\pm$0.4} \\
32 & 8.7{\color{gray}\scriptsize$\pm$18.7} & 3.6{\color{gray}\scriptsize$\pm$0.9} & 38.7{\color{gray}\scriptsize$\pm$32.1} & 7.9{\color{gray}\scriptsize$\pm$2.0} & 47.5{\color{gray}\scriptsize$\pm$35.8} & 1.8{\color{gray}\scriptsize$\pm$0.8} & 97.7{\color{gray}\scriptsize$\pm$6.7} & 2.0{\color{gray}\scriptsize$\pm$0.8} \\
\bottomrule
\end{tabular}
\end{adjustbox}
\end{table}

For \ours{}, \textit{ordered completion} increases from $81.0\%$ at $K=4$ to
$97.7\%$ at $K=32$, while time increases from $0.5$ to $2.0$~ms.
The completion gains become smaller at larger $K$ even as computation
continues to increase. The baselines do not show the same monotonic
completion trend. More stored realizations therefore accompany improved
completion for \ours{} in this sweep, but do not produce a uniform benefit
across the evaluated methods.

\textbf{View resolution.}
Table~\ref{tab:rq1_view_resolution} fixes the number of supplied views and
averages \textit{skill-context diversity}, \textit{waypoint gaps}, and all configurations at
that resolution.

\begin{table}[!h]
\centering
\caption{\textit{Ordered completion} and materialization and prediction time by supplied \textit{view resolution} $V$. Both error terms describe variation across configuration means.}
\vskip 0.1in
\label{tab:rq1_view_resolution}
\renewcommand{\arraystretch}{1.15}
\setlength{\tabcolsep}{4pt}
\begingroup
\footnotesize
\renewcommand{\arraystretch}{1.15}
\setlength{\tabcolsep}{3pt}
\begin{tabular*}{\textwidth}{@{\extracolsep{\fill}\hspace{\tabcolsep}}ccccc@{\hspace{\tabcolsep}}}
\toprule
\multicolumn{3}{l}{\textbf{Experiment Config}} & \multicolumn{2}{l}{} \\
\midrule
$K^{\dagger}$ & $H^{\dagger}$ & Views & $N_{\mathrm{cfg}}$ & $N_{\mathrm{ep}}$ \\
\midrule
$\{4,8,16,32\}$ & $\{1,\ldots,7\}$ & \makecell[c]{$V=3$\\$1$ configuration} & $4\times1\times7$ & $28\mathrm{k}$ \\
$\{4,8,16,32\}$ & $\{1,\ldots,7\}$ & \makecell[c]{$V=4$\\$2$ configurations$^{\dagger}$} & $4\times2\times7$ & $56\mathrm{k}$ \\
$\{4,8,16,32\}$ & $\{1,\ldots,7\}$ & \makecell[c]{$V=5$\\$3$ configurations$^{\dagger}$} & $4\times3\times7$ & $84\mathrm{k}$ \\
$\{4,8,16,32\}$ & $\{1,\ldots,7\}$ & \makecell[c]{$V=8$\\$1$ configuration} & $4\times1\times7$ & $28\mathrm{k}$ \\
\bottomrule
\end{tabular*}
\endgroup
\par\smallskip
\begin{adjustbox}{width=\textwidth}
\begin{tabular}{c|cc|cc|cc|>{\columncolor{gray!25}}c>{\columncolor{gray!25}}c}
\toprule
\multicolumn{3}{l}{\textbf{Results}} & \multicolumn{6}{l}{} \\
\midrule
& \multicolumn{2}{c|}{\makecell[c]{\texttt{Skill Retrieval}\\(\texttt{SR})}}
& \multicolumn{2}{c|}{\makecell[c]{\texttt{Skill Sequence}\\\texttt{Retrieval} (\texttt{SSR})}}
& \multicolumn{2}{c|}{\makecell[c]{\texttt{Skill Chaining}\\(\texttt{SC})}}
& \multicolumn{2}{c}{\makecell[c]{\ours{}\\(ours)}} \\
\cmidrule(lr){2-3}\cmidrule(lr){4-5}\cmidrule(lr){6-7}\cmidrule(lr){8-9}
$V$
& \textsc{OC}$\uparrow${\scriptsize(\%)} & Time$\downarrow${\scriptsize(ms)}
& \textsc{OC}$\uparrow${\scriptsize(\%)} & Time$\downarrow${\scriptsize(ms)}
& \textsc{OC}$\uparrow${\scriptsize(\%)} & Time$\downarrow${\scriptsize(ms)}
& {\cellcolor{white!25}\textsc{OC}$\uparrow${\scriptsize(\%)}} & {\cellcolor{white!25}Time$\downarrow${\scriptsize(ms)}} \\
\midrule
3 & 7.4{\color{gray}\scriptsize$\pm$16.5} & 1.7{\color{gray}\scriptsize$\pm$0.6} & 18.5{\color{gray}\scriptsize$\pm$28.6} & 5.0{\color{gray}\scriptsize$\pm$1.0} & 40.8{\color{gray}\scriptsize$\pm$26.9} & 0.7{\color{gray}\scriptsize$\pm$0.5} & 73.5{\color{gray}\scriptsize$\pm$21.2} & 0.8{\color{gray}\scriptsize$\pm$0.5} \\
4 & 8.7{\color{gray}\scriptsize$\pm$15.9} & 2.1{\color{gray}\scriptsize$\pm$0.8} & 27.1{\color{gray}\scriptsize$\pm$29.7} & 5.8{\color{gray}\scriptsize$\pm$1.2} & 40.1{\color{gray}\scriptsize$\pm$30.9} & 0.9{\color{gray}\scriptsize$\pm$0.6} & 90.6{\color{gray}\scriptsize$\pm$17.5} & 0.9{\color{gray}\scriptsize$\pm$0.6} \\
5 & 8.4{\color{gray}\scriptsize$\pm$15.9} & 2.5{\color{gray}\scriptsize$\pm$0.9} & 25.3{\color{gray}\scriptsize$\pm$29.6} & 6.8{\color{gray}\scriptsize$\pm$1.6} & 49.3{\color{gray}\scriptsize$\pm$31.4} & 1.0{\color{gray}\scriptsize$\pm$0.7} & 96.3{\color{gray}\scriptsize$\pm$9.4} & 1.0{\color{gray}\scriptsize$\pm$0.7} \\
8 & 7.0{\color{gray}\scriptsize$\pm$16.1} & 3.5{\color{gray}\scriptsize$\pm$1.1} & 24.3{\color{gray}\scriptsize$\pm$30.4} & 9.7{\color{gray}\scriptsize$\pm$2.1} & 93.9{\color{gray}\scriptsize$\pm$22.8} & 1.2{\color{gray}\scriptsize$\pm$1.0} & 100.0{\color{gray}\scriptsize$\pm$0.0} & 1.3{\color{gray}\scriptsize$\pm$1.0} \\
\bottomrule
\end{tabular}
\end{adjustbox}
\end{table}

\ours{} rises from $73.5\%$ \textit{ordered completion} at $V=3$ to $100.0\%$
at the supplied $V=8$ configuration. Its corresponding time increases
from $0.8$ to $1.3$~ms. \texttt{SC} also reaches its highest marginal
completion at $V=8$, whereas neither retrieval baseline improves
monotonically with view count. These trends concern the supplied
configurations, in which view count and partition structure vary together.

\subsubsection{Task-Record Sparsity Evaluation}
\label{app:task-record-sparsity}

We next fix the \textit{waypoint gap} and average \textit{procedure support} under
Section~\ref{app:evaluation-protocol}. Increasing $H$ leaves fewer
intermediate task waypoints while preserving the held-out target and
procedure records. Table~\ref{tab:rq2_waypoint_gap} reports the resulting
\textit{ordered completion} and materialization and prediction time. Its
ordered-completion deviations describe heterogeneity across the $28$
configurations at each gap.

\begin{table}[!h]
\centering
\caption{\textit{Ordered completion} and materialization and prediction time by \textit{waypoint gap} $H$. Both error terms describe variation across configuration means.}
\vskip 0.1in
\label{tab:rq2_waypoint_gap}
\renewcommand{\arraystretch}{1.15}
\setlength{\tabcolsep}{4pt}
\begingroup
\footnotesize
\renewcommand{\arraystretch}{1.15}
\setlength{\tabcolsep}{3pt}
\begin{tabular*}{\textwidth}{@{\extracolsep{\fill}\hspace{\tabcolsep}}ccccc@{\hspace{\tabcolsep}}}
\toprule
\multicolumn{3}{l}{\textbf{Experiment Config}} & \multicolumn{2}{l}{} \\
\midrule
$K^{\dagger}$ & $H$ & Views$^{\dagger}$ & $N_{\mathrm{cfg}}$ & $N_{\mathrm{ep}}$ \\
\midrule
$\{4,8,16,32\}$ & $\{1,\ldots,7\}$ & \makecell[c]{$V\in\{3,4,5,8\}$\\$7$ configurations total} & $4\times7\times7$ & $196\mathrm{k}$ \\
\bottomrule
\end{tabular*}
\endgroup
\par\smallskip
\begin{adjustbox}{width=\textwidth}
\begin{tabular}{c|cc|cc|cc|>{\columncolor{gray!25}}c>{\columncolor{gray!25}}c}
\toprule
\multicolumn{3}{l}{\textbf{Results}} & \multicolumn{6}{l}{} \\
\midrule
& \multicolumn{2}{c|}{\makecell[c]{\texttt{Skill Retrieval}\\(\texttt{SR})}}
& \multicolumn{2}{c|}{\makecell[c]{\texttt{Skill Sequence}\\\texttt{Retrieval} (\texttt{SSR})}}
& \multicolumn{2}{c|}{\makecell[c]{\texttt{Skill Chaining}\\(\texttt{SC})}}
& \multicolumn{2}{c}{\makecell[c]{\ours{}\\(ours)}} \\
\cmidrule(lr){2-3}\cmidrule(lr){4-5}\cmidrule(lr){6-7}\cmidrule(lr){8-9}
$H$
& \textsc{OC}$\uparrow${\scriptsize(\%)} & Time$\downarrow${\scriptsize(ms)}
& \textsc{OC}$\uparrow${\scriptsize(\%)} & Time$\downarrow${\scriptsize(ms)}
& \textsc{OC}$\uparrow${\scriptsize(\%)} & Time$\downarrow${\scriptsize(ms)}
& {\cellcolor{white!25}\textsc{OC}$\uparrow${\scriptsize(\%)}} & {\cellcolor{white!25}Time$\downarrow${\scriptsize(ms)}} \\
\midrule
1 & 43.8{\color{gray}\scriptsize$\pm$5.7} & 1.7{\color{gray}\scriptsize$\pm$0.6} & 68.7{\color{gray}\scriptsize$\pm$17.2} & 8.1{\color{gray}\scriptsize$\pm$2.0} & 36.2{\color{gray}\scriptsize$\pm$35.1} & 1.6{\color{gray}\scriptsize$\pm$1.1} & 92.3{\color{gray}\scriptsize$\pm$15.8} & 1.8{\color{gray}\scriptsize$\pm$1.1} \\
2 & 6.6{\color{gray}\scriptsize$\pm$12.0} & 2.0{\color{gray}\scriptsize$\pm$0.8} & 46.1{\color{gray}\scriptsize$\pm$29.3} & 7.5{\color{gray}\scriptsize$\pm$1.9} & 51.5{\color{gray}\scriptsize$\pm$34.2} & 1.1{\color{gray}\scriptsize$\pm$0.7} & 89.8{\color{gray}\scriptsize$\pm$17.8} & 1.2{\color{gray}\scriptsize$\pm$0.7} \\
3 & 0.0{\color{gray}\scriptsize$\pm$0.0} & 2.6{\color{gray}\scriptsize$\pm$1.0} & 37.6{\color{gray}\scriptsize$\pm$19.8} & 7.2{\color{gray}\scriptsize$\pm$1.8} & 53.6{\color{gray}\scriptsize$\pm$27.1} & 1.0{\color{gray}\scriptsize$\pm$0.6} & 89.3{\color{gray}\scriptsize$\pm$19.3} & 1.0{\color{gray}\scriptsize$\pm$0.6} \\
4 & 0.0{\color{gray}\scriptsize$\pm$0.0} & 2.7{\color{gray}\scriptsize$\pm$1.0} & 9.1{\color{gray}\scriptsize$\pm$16.1} & 5.4{\color{gray}\scriptsize$\pm$1.3} & 58.7{\color{gray}\scriptsize$\pm$26.8} & 0.7{\color{gray}\scriptsize$\pm$0.5} & 89.8{\color{gray}\scriptsize$\pm$17.8} & 0.8{\color{gray}\scriptsize$\pm$0.5} \\
5 & 3.2{\color{gray}\scriptsize$\pm$6.3} & 2.5{\color{gray}\scriptsize$\pm$1.0} & 6.6{\color{gray}\scriptsize$\pm$11.4} & 6.1{\color{gray}\scriptsize$\pm$2.0} & 53.6{\color{gray}\scriptsize$\pm$32.9} & 0.8{\color{gray}\scriptsize$\pm$0.5} & 91.8{\color{gray}\scriptsize$\pm$14.8} & 0.8{\color{gray}\scriptsize$\pm$0.5} \\
6 & 3.5{\color{gray}\scriptsize$\pm$6.2} & 2.5{\color{gray}\scriptsize$\pm$1.1} & 1.3{\color{gray}\scriptsize$\pm$4.1} & 6.7{\color{gray}\scriptsize$\pm$2.1} & 39.3{\color{gray}\scriptsize$\pm$29.8} & 0.8{\color{gray}\scriptsize$\pm$0.5} & 95.4{\color{gray}\scriptsize$\pm$12.3} & 0.8{\color{gray}\scriptsize$\pm$0.5} \\
7 & 0.0{\color{gray}\scriptsize$\pm$0.0} & 2.7{\color{gray}\scriptsize$\pm$1.0} & 3.5{\color{gray}\scriptsize$\pm$7.2} & 5.5{\color{gray}\scriptsize$\pm$1.7} & 69.9{\color{gray}\scriptsize$\pm$43.0} & 0.6{\color{gray}\scriptsize$\pm$0.3} & 94.9{\color{gray}\scriptsize$\pm$13.0} & 0.7{\color{gray}\scriptsize$\pm$0.4} \\
\bottomrule
\end{tabular}
\end{adjustbox}
\end{table}

\newpage
\subsubsection{Procedure Knowledge Base Construction}
\label{app:procedure-kb-cost}

To assess how procedure knowledge base preparation scales with skill-context diversity, we measure its one-time construction cost in Table~\ref{tab:rq2_procedure_build}.
This measurement includes loading the procedure records and constructing the representation used by each method.
For \ours{}, it includes constructing the marked records and procedure substrate.
We reuse the constructed knowledge base across task evaluations.

\begin{table}[!h]
\centering
\caption{Mean procedure knowledge base preparation time by skill-context diversity, in ms. Each value averages five warm construction measurements for each of seven view configurations, excluding the first construction in each process.}
\vskip 0.1in
\label{tab:rq2_procedure_build}
\small
\renewcommand{\arraystretch}{1.2}
\setlength{\tabcolsep}{3pt}
\begingroup
\footnotesize
\renewcommand{\arraystretch}{1.15}
\setlength{\tabcolsep}{3pt}
\begin{tabular*}{\textwidth}{@{\extracolsep{\fill}\hspace{\tabcolsep}}ccccc@{\hspace{\tabcolsep}}}
\toprule
\multicolumn{3}{l}{\textbf{Experiment Config}} & \multicolumn{2}{l}{} \\
\midrule
$K$ & $H$ & Views$^{\dagger}$ & $N_{\mathrm{cfg}}$ & $N_{\mathrm{ep}}$ \\
\midrule
$\{4,8,16,32\}$ & -- & \makecell[c]{$V\in\{3,4,5,8\}$\\$7$ configurations total} & $4\times7$ & -- \\
\bottomrule
\end{tabular*}
\endgroup
\par\smallskip
\begin{tabular*}{\textwidth}{@{\extracolsep{\fill}\hspace{\tabcolsep}}r|rrrr@{\hspace{\tabcolsep}}}
\toprule
\multicolumn{3}{l}{\textbf{Results}} & \multicolumn{2}{l}{} \\
\midrule
$K$ & \texttt{SR} & \texttt{SSR} & \texttt{SC} & {\ours{}} \\
\midrule
4 & 2.7 & 10.3 & 18.6 & 22.1 \\
8 & 5.3 & 21.6 & 33.5 & 33.5 \\
16 & 11.6 & 38.5 & 64.8 & 63.6 \\
32 & 24.4 & 85.4 & 140.4 & 138.0 \\
\bottomrule
\end{tabular*}
\end{table}

Preparation time increases with skill-context diversity for every method.
For \ours{}, it ranges from $22.1$ to $138.0$~ms across the evaluated settings.
The first construction in each process additionally takes $1.22$~s for \ours{} and $1.06$~s for \texttt{SC}, mostly for accelerator-library initialization, and $13$~ms and $25$~ms for \texttt{SR} and \texttt{SSR}; it is excluded from Table~\ref{tab:rq2_procedure_build}.
This preparation cost is excluded from the task-side computation reported in Section~\ref{app:evaluation-protocol}.

\subsection{Real-World Evaluation}
\label{app:realworld}

\begin{figure}[!h]
    \centering
    \includegraphics[width=1\linewidth]{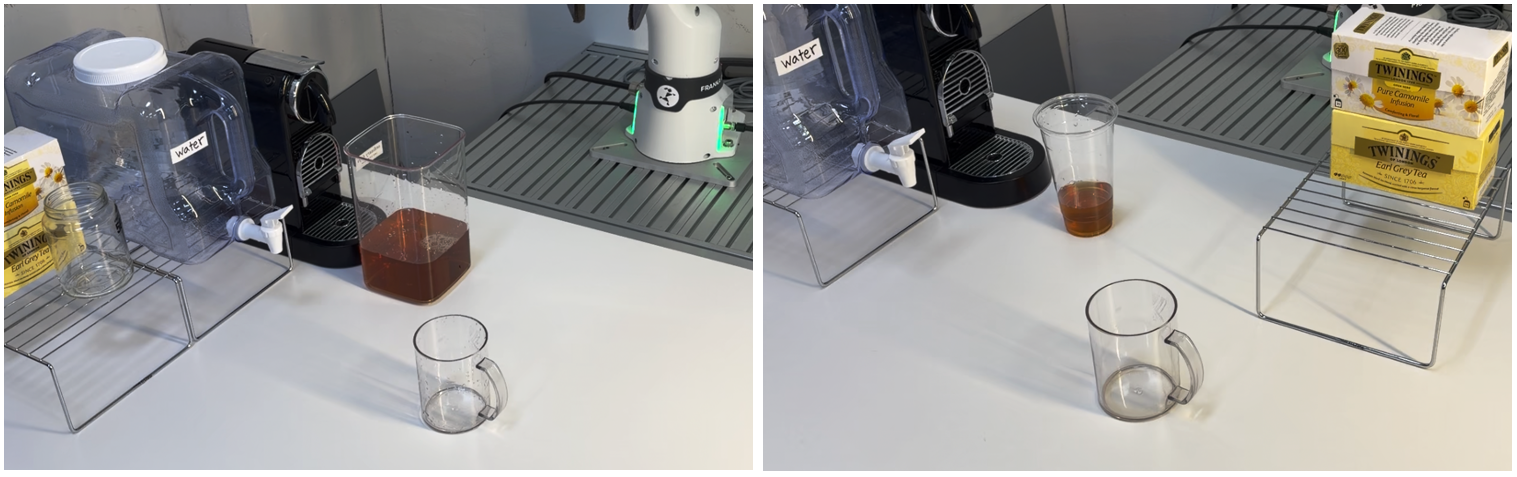}
    \vskip -0.1in
    \caption{The two physical environments used for real-world evaluation, sharing the goal \emph{pour coffee} but differing in scene composition. \textbf{(\emph{left})} Environment 1: a large source whose volume exceeds the target cup capacity. \textbf{(\emph{right})} Environment 2: a source whose volume matches the target cup capacity.}
    \vskip -0.1in
    \label{fig:realworld-envs}
\end{figure}

\textbf{Workspace configuration.} We evaluate \ours{} in a cafe workspace centered on coffee preparation. We deploy across two environments that share the goal (\emph{pour coffee}) but differ in scene composition (Figure~\ref{fig:realworld-envs}), so that procedure records collected under one scene composition are reused under another.

\textbf{Robots.} The workspace is equipped with a 7-DoF Franka Emika Research 3 tabletop manipulator.

\textbf{Perception.} The workspace is equipped with two cameras serving distinct roles. An Intel RealSense D435 RGB-D camera mounted top-down provides object-level grounding: RGB frames are passed to an object detector that returns category labels and bounding boxes for task-relevant objects, while the aligned depth channel lifts each detection to a 3D position expressed in the arm's base frame via the camera-to-robot extrinsics. A separate front-view RGB camera supplies the visual input read by the partial procedure views $\mathcal{V}$, on which the projection maps $\alpha_v$ are defined.

\textbf{Low-level controller.} Skill executions are realized as Cartesian waypoint sequences in the arm's base frame and dispatched through MoveIt, which handles inverse kinematics and trajectory planning to the underlying joint controllers.

\subsubsection{Reconfiguration Scenarios}\label{app:realworld-scenarios}

{
Both environments share the goal \emph{pour coffee}. They differ in the
source-to-target capacity relation and in the procedure records available at
deployment. The supplied views represent the source and target factors needed
for the pouring and fill-level transitions. We compare how local updates to
$\mathcal{K}_\Omega$ affect \ours{} and \texttt{SC} under the same memory.

\textbf{Case 1: interference from other views.} The source contains more liquid than
the target can hold. The task therefore requires a stopping condition before
the source becomes empty. The default views expose the stored pouring
transition through the source state but do not distinguish the target fill
level. We add a supplied target-fill view and a procedure record for the
stopping transition. \ours{} then composes the pouring and stopping skills from
their view-local effects. \texttt{SC} reuses the pouring skill from endpoint
compatibility alone and overfills the target.

\textbf{Case 2: recovery through records.} The source volume matches the target
capacity. The updated pouring segment requires one skill ($L_h{=}1$), while the
full task contains two ordered skills. The memory also contains endpoint-compatible
records whose target-side effects do not match the requested fill transition.
\texttt{SC} can therefore select an incorrect record from endpoint compatibility.
We append the correct pouring record without changing the skill library. Its
target-view transition matches the requested effect, so \ours{} binds the
correct pouring skill within the two-skill task.
}

{
\subsubsection{Record-Appending Scenario}
For the second scenario above, procedure records are collected under three
scene conditions. These collection conditions are distinct from the two
physical reconfiguration scenarios. The evaluation combines background and
manipulated-object conditions not paired during collection. We run ten trials
per method under the same lighting and nominal object placement.
Sequence success requires the ordered two-skill sequence from the eight-skill
library. Task success additionally requires completing the pour without
spilling. Over the ten trials per method, \ours{} achieves $10/10$ sequence success and $7/10$ task success, whereas \texttt{SC} achieves $0/10$ on both.

The
three failed \ours{} trials use the correct materialized sequence and are
attributed to grasp, alignment, or depth-localization errors.
}

\section{Additional Experiments}

\textbf{Supplementary evaluation.}
\label{app:supplementary-protocol}
To evaluate supplementary sensitivity, we use the experiment-specific configurations listed with each result table.
Evaluation and aggregation follow Section~\ref{app:evaluation-protocol}.

\subsection{Ablations}
\label{app:component-selection}

Table~\ref{tab:salience-ablation} removes one kind of evidence of Appendix~\ref{app:procedure_kb_details} at a time from task materialization, with the supplied views fixed. We use the sweep in Table~\ref{tab:shared-evaluation-conditions} and the aggregation in Section~\ref{app:evaluation-protocol}. Both axes leave the decision-point refinement of Appendix~\ref{app:task_materialization_details} unchanged.
\begin{itemize}[leftmargin=*]
    \item \textbf{View selection.} Candidate-specific selection uses the identification evidence to choose the mask $W_{\omega,h}$ and the candidates of $\Omega_h^0$, as specified in Appendix~\ref{app:task_materialization_details}. Uniform selection removes this evidence from segment binding. It sets $W_{\omega,h}=\mathds{1}_h$ for every candidate and keeps every candidate when no signature covers $\mathcal{V}_h$.
    \item \textbf{Chaining prior.} Both settings keep the path objective of Eq.~\eqref{eq:app-path-objective}. \emph{Prior-guided chaining} reorders the composed chain with the empirical skill adjacency $\rightarrow_\Omega$ when it admits exactly one order covering the chained skills. \emph{Chaining} keeps the composed order and uses no skill order of the procedure records.
\end{itemize}
All other \ours{} settings are the same. The skill order that $\rightarrow_\Omega$ reads is the same evidence that \texttt{SSR} retrieves, and the chaining setting uses no more order evidence than \texttt{SC}.

We also include \texttt{SC}, which shares the decision-point refinement of \ours{} and orders its candidates by the same path objective over the centroids. Its candidate selection differs from the uniform chaining setting of \ours{}. \texttt{SC} does not restrict the candidates by signature. It keeps the condition that the dominant view of a candidate is active. It keeps every candidate with positive support instead of applying $\varepsilon_\rho$.

\begin{table}[!h]
\centering
\small
\renewcommand{\arraystretch}{1.15}
\setlength{\tabcolsep}{3pt}
\caption{Mean OC for the view-selection and chaining-prior ablations of \ours{} and the \texttt{SC} comparison on the sweep in Table~\ref{tab:shared-evaluation-conditions}. Configuration and episode counts are per row. The default \ours{} result is bold.}
\vskip 0.1in
\label{tab:salience-ablation}
\label{tab:component-factorial}
\begingroup
\footnotesize
\begin{tabular*}{\textwidth}{@{\extracolsep{\fill}\hspace{\tabcolsep}}ccccc@{\hspace{\tabcolsep}}}
\toprule
\multicolumn{3}{l}{\textbf{Experiment Config}} & \multicolumn{2}{l}{} \\
\midrule
$K^{\dagger}$ & $H^{\dagger}$ & Views$^{\dagger}$ & $N_{\mathrm{cfg}}$ & $N_{\mathrm{ep}}$ \\
\midrule
$\{4,8,16,32\}$ & $\{1,\ldots,7\}$ & \makecell[c]{$V\in\{3,4,5,8\}$\\$7$ configurations total} & $4\times7\times7$ & $196\mathrm{k}$ \\
\bottomrule
\end{tabular*}
\endgroup
\par\smallskip
\begin{tabular*}{\textwidth}{@{\extracolsep{\fill}\hspace{\tabcolsep}}llcc@{\hspace{\tabcolsep}}}
\toprule
\multicolumn{4}{l}{\textbf{Results}} \\
\midrule
Method & View selection & Chaining prior & \textsc{OC}$\uparrow${\scriptsize(\%)} \\
\midrule
\multirow{4}{*}{\ours{}} & \multirow{2}{*}{Candidate-specific} & Prior-guided chaining & \textbf{91.9} \\
 & & Chaining & 86.3 \\
\cmidrule(lr){2-4}
 & \multirow{2}{*}{Uniform} & Prior-guided chaining & 89.1 \\
 & & Chaining & 70.6 \\
\midrule
\texttt{SC} & -- & -- & 51.8 \\
\bottomrule
\end{tabular*}
\end{table}

Removing either kind of evidence lowers OC. Without the chaining prior, uniform selection lowers OC from $86.3\%$ to $70.6\%$, while with the prior it lowers OC only from $91.9\%$ to $89.1\%$.

{
\subsection{Analysis}
\subsubsection{Robustness to Procedure-View Specification}
\label{app:view-robustness}

To assess sensitivity to procedure-view specification, we vary partition structure and overlap under the supplementary protocol in Section~\ref{app:supplementary-protocol}.
Both reference and random assignments use separate view-construction seeds while keeping the underlying observations fixed.
The partitioning seeds for random assignments are distinct from the evaluation seeds.
This reference distribution differs from the fixed views used in the main evaluation.

To examine the role of the supplied partition structure, we redistribute the same observation dimensions into random partitions with the same view count.
Table~\ref{tab:random-view-partitions} compares these partitions with the reference object-factor assignments.

\begin{table}[!h]
\centering
\small
\renewcommand{\arraystretch}{1.15}
\setlength{\tabcolsep}{3pt}
\caption{OC under reference and random procedure-view partitions. $S_p$ denotes the number of view-construction seeds, distinct from evaluation seeds. For random partitions, each seed determines the assignment of observation dimensions to views. Under random partitions, \ours{} tuned canonicalization for random partitions.}
\vskip 0.1in
\label{tab:random-view-partitions}
\begingroup
\footnotesize
\renewcommand{\arraystretch}{1.15}
\setlength{\tabcolsep}{3pt}
\begin{tabular*}{\textwidth}{@{\extracolsep{\fill}\hspace{\tabcolsep}}cccccc@{\hspace{\tabcolsep}}}
\toprule
\multicolumn{3}{l}{\textbf{Experiment Config}} & \multicolumn{3}{l}{} \\
\midrule
$K^{\dagger}$ & $H^{\dagger}$ & Views$^{\dagger}$ & $S_p^{\dagger}$ & $N_{\mathrm{cfg}}$ & $N_{\mathrm{ep}}$ \\
\midrule
$\{4,8,16,32\}$ & $\{1,2,4,7\}$ & $V\in\{3,4,5,8\}$ & $5$ & $4\times4\times4\times5$ & $320\mathrm{k}$ \\
\bottomrule
\end{tabular*}
\endgroup
\par\smallskip
\begin{tabular*}{\textwidth}{@{\extracolsep{\fill}\hspace{\tabcolsep}}l|rr@{\hspace{\tabcolsep}}}
\toprule
\multicolumn{3}{l}{\textbf{Results}} \\
\midrule
Method & Reference views & Random partitions \\
\midrule
\texttt{SR} & $12.5\,{\scriptsize\pm 0.2}$ & $13.1\,{\scriptsize\pm 1.1}$ \\
\texttt{SSR} & $31.2\,{\scriptsize\pm 0.5}$ & $21.3\,{\scriptsize\pm 1.1}$ \\
\texttt{SC} & $55.2\,{\scriptsize\pm 1.9}$ & $34.9\,{\scriptsize\pm 6.4}$ \\
\midrule
\ours{} & $90.5\,{\scriptsize\pm 2.4}$ & $63.2\,{\scriptsize\pm 8.3}$ \\
\bottomrule
\end{tabular*}
\end{table}

Random partitions lower the OC of \ours{}, \texttt{SC}, and \texttt{SSR} and widen the variation across construction seeds, while \texttt{SR} remains near its low reference level.
\ours{} retains the highest OC under both assignments.
The methods that compose skills over the supplied views thus depend on how the same observation dimensions are assigned to views.

To measure sensitivity to overlapping views, we copy a varying fraction of factors into another existing view, as reported in Table~\ref{tab:factor-duplicated-views}.
The number of views, underlying observations, and remaining evaluation settings are unchanged.
At duplication fraction $1$, every factor appears in two views.

\begin{table}[!h]
\centering
\small
\renewcommand{\arraystretch}{1.15}
\setlength{\tabcolsep}{3pt}
\caption{OC as the factor-duplication fraction varies. $S_p$ denotes the number of view-construction seeds, distinct from evaluation seeds.}
\vskip 0.1in
\label{tab:factor-duplicated-views}
\begingroup
\footnotesize
\renewcommand{\arraystretch}{1.15}
\setlength{\tabcolsep}{3pt}
\begin{tabular*}{\textwidth}{@{\extracolsep{\fill}\hspace{\tabcolsep}}cccccc@{\hspace{\tabcolsep}}}
\toprule
\multicolumn{3}{l}{\textbf{Experiment Config}} & \multicolumn{3}{l}{} \\
\midrule
$K^{\dagger}$ & $H^{\dagger}$ & Views$^{\dagger}$ & $S_p^{\dagger}$ & $N_{\mathrm{cfg}}$ & $N_{\mathrm{ep}}$ \\
\midrule
$\{4,8,16,32\}$ & $\{1,2,4,7\}$ & $V\in\{3,4,5,8\}$ & $5$ & $4\times4\times4\times5$ & $320\mathrm{k}$ \\
\bottomrule
\end{tabular*}
\endgroup
\par\smallskip
\begin{tabular*}{\textwidth}{@{\extracolsep{\fill}\hspace{\tabcolsep}}l|rrrr@{\hspace{\tabcolsep}}}
\toprule
\multicolumn{3}{l}{\textbf{Results}} & \multicolumn{2}{l}{} \\
\midrule
Duplication fraction & 0 & 0.25 & 0.5 & 1.0 \\
\midrule
\texttt{SR} & $12.5\,{\scriptsize\pm 0.2}$ & $14.7\,{\scriptsize\pm 1.6}$ & $14.7\,{\scriptsize\pm 2.9}$ & $11.6\,{\scriptsize\pm 0.5}$ \\
\texttt{SSR} & $31.2\,{\scriptsize\pm 0.5}$ & $36.0\,{\scriptsize\pm 10.8}$ & $35.2\,{\scriptsize\pm 8.7}$ & $27.3\,{\scriptsize\pm 2.7}$ \\
\texttt{SC} & $55.2\,{\scriptsize\pm 1.9}$ & $49.1\,{\scriptsize\pm 7.9}$ & $47.7\,{\scriptsize\pm 5.5}$ & $33.6\,{\scriptsize\pm 4.7}$ \\
\midrule
\ours{} & $90.5\,{\scriptsize\pm 2.4}$ & $68.0\,{\scriptsize\pm 12.4}$ & $65.7\,{\scriptsize\pm 8.5}$ & $49.7\,{\scriptsize\pm 13.1}$ \\
\bottomrule
\end{tabular*}
\end{table}

As the duplication fraction increases, the OC of \ours{} and \texttt{SC} decreases, with the largest drop when every factor is duplicated.
\texttt{SSR} and \texttt{SR} vary within a narrower range around their low reference levels.
\ours{} retains the highest OC at every duplication fraction.
Overlapping views therefore degrade the methods that compose skills through view-level structure, and duplicating observations across views does not preserve the performance of the reference interface.
}

\subsubsection{Canonicalization Sensitivity}
\label{app:canonicalization-sensitivity}

To assess how endpoint canonicalization affects completion, we vary the canonicalization map while holding the remaining materialization settings fixed on the sweep in Table~\ref{tab:shared-evaluation-conditions}.
Table~\ref{tab:canonicalization-alternatives} compares DBSCAN with connected components of an $\epsilon$-neighborhood graph and KMeans with its cluster count selected by the silhouette score.
The neighborhood radius is $0.1$ for DBSCAN and the $\epsilon$-neighborhood graph.

\begin{table}[!h]
\centering
\small
\renewcommand{\arraystretch}{1.25}
\setlength{\tabcolsep}{6pt}
\caption{\ours{} OC across endpoint canonicalization procedures.}
\vskip 0.1in
\label{tab:canonicalization-alternatives}
\begingroup
\footnotesize
\renewcommand{\arraystretch}{1.15}
\setlength{\tabcolsep}{3pt}
\begin{tabular*}{\textwidth}{@{\extracolsep{\fill}\hspace{\tabcolsep}}ccccc@{\hspace{\tabcolsep}}}
\toprule
\multicolumn{3}{l}{\textbf{Experiment Config}} & \multicolumn{2}{l}{} \\
\midrule
$K^{\dagger}$ & $H^{\dagger}$ & Views$^{\dagger}$ & $N_{\mathrm{cfg}}$ & $N_{\mathrm{ep}}$ \\
\midrule
$\{4,8,16,32\}$ & $\{1,\ldots,7\}$ & \makecell[c]{$V\in\{3,4,5,8\}$\\$7$ configurations total} & $4\times7\times7$ & $196\mathrm{k}$ \\
\bottomrule
\end{tabular*}
\endgroup
\par\smallskip
\begin{tabular*}{\textwidth}{@{\extracolsep{\fill}\hspace{\tabcolsep}}lccc@{\hspace{\tabcolsep}}}
\toprule
\multicolumn{3}{l}{\textbf{Results}} & \multicolumn{1}{l}{} \\
\midrule
\multirow{2}{*}{Metric} & \multicolumn{3}{c}{Canonicalization algorithm} \\
\cmidrule(lr){2-4}
& DBSCAN & $\epsilon$-neighborhood graph & Silhouette KMeans \\
\midrule
\textsc{OC}$\uparrow${\scriptsize(\%)} & 91.9 & 88.0 & 88.3 \\
\bottomrule
\end{tabular*}
\end{table}

Mean OC ranges from $88.0\%$ to $91.9\%$ across the three procedures.
DBSCAN attains the highest mean in this comparison.

To assess sensitivity to DBSCAN parameters, we evaluate all $54$ combinations of nine neighborhood radii and six minimum sample counts in Table~\ref{tab:dbscan-sensitivity}.
The default remains radius $0.10$ with minimum sample count $20$.

\begin{table}[!h]
\centering
\small
\renewcommand{\arraystretch}{1.25}
\setlength{\tabcolsep}{6pt}
\caption{\ours{} OC across DBSCAN parameters. The bold entry denotes the default setting.}
\vskip 0.1in
\label{tab:dbscan-sensitivity}
\begingroup
\footnotesize
\renewcommand{\arraystretch}{1.15}
\setlength{\tabcolsep}{3pt}
\begin{tabular*}{\textwidth}{@{\extracolsep{\fill}\hspace{\tabcolsep}}ccccc@{\hspace{\tabcolsep}}}
\toprule
\multicolumn{3}{l}{\textbf{Experiment Config}} & \multicolumn{2}{l}{} \\
\midrule
$K^{\dagger}$ & $H^{\dagger}$ & Views$^{\dagger}$ & $N_{\mathrm{cfg}}$ & $N_{\mathrm{ep}}$ \\
\midrule
$\{4,8,16,32\}$ & $\{1,\ldots,7\}$ & \makecell[c]{$V\in\{3,4,5,8\}$\\$7$ configurations total} & $4\times7\times7$ & $196\mathrm{k}$ \\
\bottomrule
\end{tabular*}
\endgroup
\par\smallskip
\begin{tabular*}{\textwidth}{@{\extracolsep{\fill}\hspace{\tabcolsep}}ccccccc@{\hspace{\tabcolsep}}}
\toprule
\multicolumn{3}{l}{\textbf{Results}} & \multicolumn{4}{l}{} \\
\midrule
\multirow{2}{*}[-0.6ex]{\makecell[c]{Neighborhood\\radius $\epsilon$}} & \multicolumn{6}{c}{Minimum sample count} \\
\cmidrule(lr){2-7}
& 1 & 5 & 10 & 20 & 40 & 80 \\
\midrule
0.010 & 33.4 & 33.6 & 24.0 & 14.8 & 18.9 & 11.7 \\
0.025 & 51.8 & 68.3 & 71.1 & 42.8 & 45.6 & 25.2 \\
0.050 & 78.8 & 93.7 & 90.0 & 85.1 & 85.9 & 63.3 \\
0.075 & 88.0 & 94.1 & 93.8 & 88.3 & 83.1 & 74.5 \\
0.100 & 88.0 & 94.4 & 95.6 & \textbf{91.9} & 85.7 & 75.5 \\
0.150 & 88.0 & 94.1 & 97.2 & 96.1 & 87.8 & 79.2 \\
0.200 & 97.2 & 97.2 & 97.2 & 97.2 & 92.3 & 83.2 \\
0.300 & 14.3 & 14.3 & 14.3 & 14.3 & 14.3 & 14.3 \\
0.500 & 8.2 & 8.2 & 8.2 & 8.2 & 8.2 & 7.8 \\
\bottomrule
\end{tabular*}
\end{table}

Across the evaluated combinations, mean OC ranges from $7.8\%$ to $97.2\%$.
For radii from $0.05$ to $0.20$ and minimum sample counts from $5$ to $20$, it lies between $85.1\%$ and $97.2\%$.
Five combinations attain the maximum, while the larger radii $0.30$ and $0.50$ yield substantially lower completion.

\subsubsection{Visual Procedure Views}
\label{app:visual-procedure-views}
{
To assess whether \ours{} can compose skills from visual observations, we define one procedure view per camera using DINOv3~\citep{simeoni2025dinov3} image embeddings.
These views exclude sensor-state features.
Figure~\ref{fig:visual-procedure-views} illustrates the five camera observations of the same Kitchen state.
We vary the number of cameras from one to five, using the first $n$ cameras shown from left to right for each camera count $n$.

\begin{figure}[!ht]
    \centering
    \includegraphics[width=\linewidth]{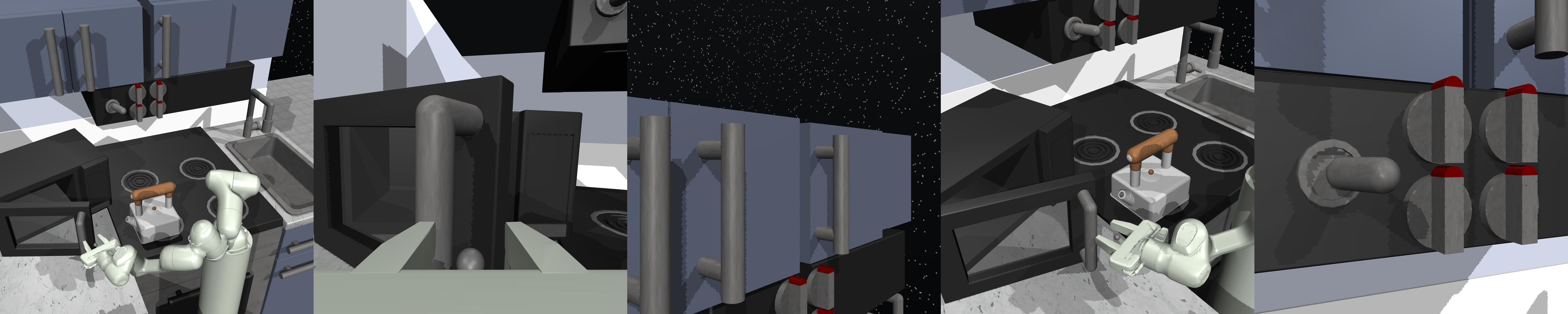}
    \par\smallskip
    {\scriptsize
    \makebox[0.2\linewidth]{(a) Overview}%
    \makebox[0.2\linewidth]{(b) Wrist}%
    \makebox[0.2\linewidth]{(c) Cabinets}%
    \makebox[0.2\linewidth]{(d) Stovetop}%
    \makebox[0.2\linewidth]{(e) Controls}\par}
    \caption{Five camera views of the same Kitchen state, rendered at higher resolution for visualization. Each image supplies a separate visual procedure view.}
    \label{fig:visual-procedure-views}
\end{figure}

To examine the effect of camera coverage, we compare the five camera subsets in Table~\ref{tab:visual-procedure-views}.
\ours{} constructs a skill sequence before execution and does not replan from new camera observations during execution.
Table~\ref{tab:visual-procedure-views} reports OC for each camera count, averaging over skill-context diversity and waypoint gap.

\begin{table}[!h]
\centering
\small
\renewcommand{\arraystretch}{1.15}
\setlength{\tabcolsep}{3pt}
\caption{OC across camera counts with visual procedure views. Mean additionally averages the camera subsets.}
\vskip 0.1in
\label{tab:visual-procedure-views}
\begingroup
\footnotesize
\renewcommand{\arraystretch}{1.15}
\setlength{\tabcolsep}{3pt}
\begin{tabular*}{\textwidth}{@{\extracolsep{\fill}\hspace{\tabcolsep}}ccccc@{\hspace{\tabcolsep}}}
\toprule
\multicolumn{3}{l}{\textbf{Experiment Config}} & \multicolumn{2}{l}{} \\
\midrule
$K^{\dagger}$ & $H^{\dagger}$ & Views & $N_{\mathrm{cfg}}$ & $N_{\mathrm{ep}}$ \\
\midrule
$\{4,8,16,32\}$ & $\{1,2,4,7\}$ & \makecell[c]{$V\in\{1,\ldots,5\}$\\One subset per $V$} & $4\times5\times4$ & $80\mathrm{k}$ \\
\bottomrule
\end{tabular*}
\endgroup
\par\smallskip
\begin{tabular*}{\textwidth}{@{\extracolsep{\fill}\hspace{\tabcolsep}}l|rrrrr|r@{\hspace{\tabcolsep}}}
\toprule
\multicolumn{3}{l}{\textbf{Results}} & \multicolumn{4}{l}{} \\
\midrule
Camera count & 1 & 2 & 3 & 4 & 5 & Mean \\
\midrule
\texttt{SR} & 11.3 & 15.2 & 16.0 & 17.4 & 9.2 & 13.8 \\
\texttt{SSR} & 19.3 & 11.7 & 13.1 & 12.4 & 14.2 & 14.1 \\
\texttt{SC} & 42.9 & 50.9 & 53.6 & 53.6 & 47.3 & 49.6 \\
\midrule
\ours{} & 82.1 & 91.1 & 85.7 & 92.9 & 83.0 & 87.0 \\
\bottomrule
\end{tabular*}
\end{table}

\ours{} attains $87.0\%$ mean OC across the five camera subsets.
The highest camera-count mean is $92.9\%$ with four cameras, and performance does not increase monotonically with camera count.
These results concern the fixed camera subsets shown above rather than an average over alternative subsets of each size.

To examine how visual composition depends on procedure support and task sparsity, we report the skill-context diversity and waypoint-gap marginals in Table~\ref{tab:visual-procedure-axes}.
For these visual results, the standard deviation across evaluation-seed means is $0.0$ percentage points for every reported marginal; this does not measure variation across configurations.

\begin{table}[!h]
\centering
\small
\renewcommand{\arraystretch}{1.15}
\setlength{\tabcolsep}{3pt}
\caption{OC by skill-context diversity and waypoint gap with visual procedure views. Both marginals average the same five camera subsets.}
\vskip 0.1in
\label{tab:visual-procedure-axes}
\begingroup
\footnotesize
\renewcommand{\arraystretch}{1.15}
\setlength{\tabcolsep}{3pt}
\begin{tabular*}{\textwidth}{@{\extracolsep{\fill}\hspace{\tabcolsep}}lccccc@{\hspace{\tabcolsep}}}
\toprule
\multicolumn{3}{l}{\textbf{Experiment Config}} & \multicolumn{3}{l}{} \\
\midrule
Report & $K$ & $H$ & Views & $N_{\mathrm{cfg}}$ & $N_{\mathrm{ep}}$ \\
\midrule
$K$ & $\{4,8,16,32\}$ & $\{1,2,4,7\}^{\dagger}$ & \makecell[c]{$V\in\{1,\ldots,5\}^{\dagger}$\\One subset per $V$} & $4\times5\times4$ & $80\mathrm{k}$ \\
$H$ & $\{4,8,16,32\}^{\dagger}$ & $\{1,2,4,7\}$ & \makecell[c]{$V\in\{1,\ldots,5\}^{\dagger}$\\One subset per $V$} & $4\times5\times4$ & $80\mathrm{k}$ \\
\bottomrule
\end{tabular*}
\endgroup
\par\smallskip
\begin{tabular*}{\textwidth}{@{\extracolsep{\fill}\hspace{\tabcolsep}}lcccccccc@{\hspace{\tabcolsep}}}
\toprule
\multicolumn{3}{l}{\textbf{Results}} & \multicolumn{6}{l}{} \\
\midrule
\multirow{2}{*}[-0.6ex]{Method} & \multicolumn{4}{c}{Skill-context diversity $K$} & \multicolumn{4}{c}{Waypoint gap $H$} \\
\cmidrule(lr){2-5}\cmidrule(lr){6-9}
& $4$ & $8$ & $16$ & $32$ & $1$ & $2$ & $4$ & $7$ \\
\midrule
\texttt{SR} & 15.9 & 11.5 & 14.2 & 13.7 & 49.4 & 0.0 & 0.0 & 5.8 \\
\texttt{SSR} & 20.8 & 9.6 & 15.3 & 10.9 & 33.4 & 7.3 & 5.8 & 10.1 \\
\texttt{SC} & 43.6 & 47.9 & 58.6 & 48.6 & 40.0 & 47.1 & 48.6 & 62.9 \\
\midrule
\ours{} & 60.0 & 95.0 & 92.9 & 100.0 & 94.3 & 91.4 & 86.4 & 75.7 \\
\bottomrule
\end{tabular*}
\end{table}

Mean OC decreases from $94.3\%$ at $H=1$ to $75.7\%$ at $H=7$.
Across diversity settings, it ranges from $60.0\%$ to $100.0\%$ without a monotonic increase.
}

\subsubsection{Prototype-based Skill Routing}

To test whether compact summaries of skill records improve retrieval, we adapt prototype-based routing~\cite{iscil} to \texttt{SR} and \texttt{SSR}.
We construct five KMeans prototypes per skill in each procedure view from record transitions and score each skill by its nearest prototype, averaging distances across views.
For \texttt{SSR}, prototype summaries also replace record-level evidence in sequence scoring, while the sequence-search procedure is unchanged.
The \texttt{SR} and \texttt{SSR} rows in Table~\ref{tab:prototype-routing} are the main-paper \texttt{SR} and \texttt{SSR} at $V=8$, while the prototype variants retain all represented skills as candidates.

\ours{} denotes the main-paper \ours{} with the supplied full-resolution procedure views ($V=8$).
For the task specification, all methods receive the same sparse waypoints.
For procedure evidence, all methods use sparse procedure-record endpoints under $V=8$ views, with the prototype variants replacing record-level evidence by the summaries defined above.
The comparison therefore measures prototype routing within each retrieval baseline, while \ours{} provides a reference under the same task inputs.

\begin{table}[!h]
\centering
\small
\renewcommand{\arraystretch}{1.15}
\setlength{\tabcolsep}{3pt}
\caption{OC across waypoint gaps for the main-paper retrieval baselines and their prototype variants at $V=8$. Mean additionally averages $H$.}
\vskip 0.1in
\label{tab:prototype-routing}
\begingroup
\footnotesize
\renewcommand{\arraystretch}{1.15}
\setlength{\tabcolsep}{3pt}
\begin{tabular*}{\textwidth}{@{\extracolsep{\fill}\hspace{\tabcolsep}}ccccc@{\hspace{\tabcolsep}}}
\toprule
\multicolumn{3}{l}{\textbf{Experiment Config}} & \multicolumn{2}{l}{} \\
\midrule
$K^{\dagger}$ & $H$ & Views & $N_{\mathrm{cfg}}$ & $N_{\mathrm{ep}}$ \\
\midrule
$\{4,8,16,32\}$ & $\{1,2,4,7\}$ & $V=8$ & $4\times1\times4$ & $16\mathrm{k}$ \\
\bottomrule
\end{tabular*}
\endgroup
\par\smallskip
\begin{tabular*}{\textwidth}{@{\extracolsep{\fill}\hspace{\tabcolsep}}lrrrrr@{\hspace{\tabcolsep}}}
\toprule
\multicolumn{3}{l}{\textbf{Results}} & \multicolumn{3}{l}{} \\
\midrule
Method & $H=1$ & $H=2$ & $H=4$ & $H=7$ & Mean \\
\midrule
\texttt{SR} & 44.5 & 4.2 & 0.0 & 0.0 & 12.2 \\

\texttt{SR}+prototype & 41.3 & 5.4 & 0.0 & 0.0 & 11.7 \\
\midrule
\texttt{SSR} & 71.2 & 47.0 & 8.5 & 3.5 & 32.5 \\

\texttt{SSR}+prototype & 98.7 & 94.4 & 28.0 & 0.0 & 55.3 \\
\midrule
\ours{} & 100.0 & 100.0 & 100.0 & 100.0 & 100.0 \\

\bottomrule
\end{tabular*}
\end{table}

Prototypes increase mean \texttt{SSR} OC from $32.5\%$ to $55.3\%$, while the corresponding \texttt{SR} means are $12.2\%$ and $11.7\%$.
The \texttt{SSR} improvement holds at $H=1,2,4$, but its prototype variant reaches $0.0\%$ at $H=7$.
The benefit therefore depends on the retrieval method and waypoint gap.

To characterize prototype retrieval across view configurations and task sparsity, we evaluate the five-prototype variants over the full sweep in Section~\ref{app:evaluation-protocol}.
Table~\ref{tab:prototype-full-grid} reports the resulting waypoint-gap marginals.

\begin{table}[!h]
\centering
\small
\renewcommand{\arraystretch}{1.15}
\setlength{\tabcolsep}{3pt}
\caption{OC of the main-paper retrieval baselines and their five-prototype variants across waypoint gaps, averaged over skill-context diversity and all supplied view configurations. Mean additionally averages $H$.}
\vskip 0.1in
\label{tab:prototype-full-grid}
\begingroup
\footnotesize
\renewcommand{\arraystretch}{1.15}
\setlength{\tabcolsep}{3pt}
\begin{tabular*}{\textwidth}{@{\extracolsep{\fill}\hspace{\tabcolsep}}ccccc@{\hspace{\tabcolsep}}}
\toprule
\multicolumn{3}{l}{\textbf{Experiment Config}} & \multicolumn{2}{l}{} \\
\midrule
$K^{\dagger}$ & $H$ & Views$^{\dagger}$ & $N_{\mathrm{cfg}}$ & $N_{\mathrm{ep}}$ \\
\midrule
$\{4,8,16,32\}$ & $\{1,\ldots,7\}$ & \makecell[c]{$V\in\{3,4,5,8\}$\\$7$ configurations total} & $4\times7\times7$ & $196\mathrm{k}$ \\
\bottomrule
\end{tabular*}
\endgroup
\par\smallskip
\begin{tabular*}{\textwidth}{@{\extracolsep{\fill}\hspace{\tabcolsep}}l|rrrrrrrr@{\hspace{\tabcolsep}}}
\toprule
\multicolumn{3}{l}{\textbf{Results}} & \multicolumn{6}{l}{} \\
\midrule
Method / $H$ & 1 & 2 & 3 & 4 & 5 & 6 & 7 & Mean \\
\midrule
\texttt{SR} & 43.8 & 6.6 & 0.0 & 0.0 & 3.2 & 3.5 & 0.0 & 8.1 \\
\texttt{SR}+prototype & 41.4 & 6.7 & 0.0 & 0.0 & 10.6 & 0.6 & 0.0 & 8.5 \\
\midrule
\texttt{SSR} & 68.7 & 46.1 & 37.6 & 9.1 & 6.6 & 1.3 & 3.5 & 24.7 \\
\texttt{SSR}+prototype & 98.1 & 90.3 & 39.0 & 27.6 & 1.5 & 5.9 & 0.0 & 37.5 \\
\midrule
\ours{} & 92.3 & 89.8 & 89.3 & 89.8 & 91.8 & 95.4 & 94.9 & 91.9 \\
\bottomrule
\end{tabular*}
\end{table}

Across all configurations, \texttt{SSR}+prototype attains higher mean OC than \texttt{SR}+prototype, with averages of $37.5\%$ and $8.5\%$, respectively.
Both reach $0.0\%$ at $H=7$, and neither exhibits a monotonic trend across all gaps.

{
\subsection{Dense Task Specifications}
\label{app:dense-comparison}
{
To compare \ours{} with composition from dense specifications, we evaluate adapted sub-trajectory retrieval and graph-search mechanisms.
Direct comparisons with the complete prior systems would change the learning or control interface.
STRAP~\cite{memmel2025strap} and graph-based imitation methods~\cite{yin2024offline,wang2025robust} use retrieved or stitched evidence for policy learning.
Goal-conditioned graph planners~\cite{opryshko2025test,baek2025graph} instead assume a goal-conditioned controller and learned value or temporal-distance representations.
Our setting provides a fixed library of executable skills without these learned planning components.
We therefore adapt their sub-trajectory retrieval and graph-search mechanisms to select skills from the existing library without policy training.

To compare these mechanisms across procedure-support conditions, we average over the KB settings listed in each condition table.
In the sub-trajectory retrieval and graph-search comparisons below, \ours{} denotes the reference evaluated with the supplied full-resolution procedure views ($V=8$) and the task inputs specified below.
Methods that use procedure views receive this configuration unless stated otherwise.

\subsubsection{Main Evaluation with Dense Task Specifications}
\label{app:dense-main-evaluation}

Tables~\ref{tab:dense-skill-context-diversity} and~\ref{tab:dense-view-resolution} report the main evaluation with dense task specifications.
To let \texttt{SR} and \texttt{SSR} exploit the dense trajectory, we additionally equip them with bisection goals.
Each goal is the best-scoring midpoint of a four-level recursive bisection of the task frames that follow the frame nearest the current observation.

With these goals, \texttt{SR} and \texttt{SSR} recover most subtasks by following the demonstrated frames, and \texttt{SR} exceeds \ours{} in OC.
Under the sparse waypoints of Tables~\ref{tab:rq1_headline_average} and~\ref{tab:rq2_waypoint_gap}, the omitted frames leave retrieval without goals to follow, whereas \ours{} attains similar OC under both specifications by composing skills from view-level changes.
This contrast locates the contribution of \ours{} in sparse governing information, where a task specification gives abstract goals rather than the behavior that realizes them.

\begin{table}[!h]
\centering
\caption{\textit{Ordered completion} and task-inference time by \textit{skill-context diversity} $K$ with dense task specifications. Overall averages all configurations. Both error terms describe variation across configuration means. $^{*}$\texttt{SR} and \texttt{SSR} with bisection goals.}
\vskip 0.1in
\label{tab:dense-skill-context-diversity}
\renewcommand{\arraystretch}{1.15}
\setlength{\tabcolsep}{4pt}
\begingroup
\footnotesize
\renewcommand{\arraystretch}{1.15}
\setlength{\tabcolsep}{3pt}
\begin{tabular*}{\textwidth}{@{\extracolsep{\fill}\hspace{\tabcolsep}}ccccc@{\hspace{\tabcolsep}}}
\toprule
\multicolumn{3}{l}{\textbf{Experiment Config}} & \multicolumn{2}{l}{} \\
\midrule
$K$ & Task input & Views$^{\dagger}$ & $N_{\mathrm{cfg}}$ & $N_{\mathrm{ep}}$ \\
\midrule
$\{4,8,16,32\}$ & Dense & \makecell[c]{$V\in\{3,4,5,8\}$\\$7$ configurations total} & $4\times7$ & $28\mathrm{k}$ \\
\bottomrule
\end{tabular*}
\endgroup
\par\smallskip
\begin{adjustbox}{width=\textwidth}
\begin{tabular}{c|cc|cc|cc|>{\columncolor{gray!25}}c>{\columncolor{gray!25}}c}
\toprule
\multicolumn{3}{l}{\textbf{Results}} & \multicolumn{6}{l}{} \\
\midrule
& \multicolumn{2}{c|}{\makecell[c]{\texttt{Skill Retrieval}\\(\texttt{SR})$^{*}$}}
& \multicolumn{2}{c|}{\makecell[c]{\texttt{Skill Sequence}\\\texttt{Retrieval} (\texttt{SSR})$^{*}$}}
& \multicolumn{2}{c|}{\makecell[c]{\texttt{Skill Chaining}\\(\texttt{SC})}}
& \multicolumn{2}{c}{\makecell[c]{\ours{}\\(ours)}} \\
\cmidrule(lr){2-3}\cmidrule(lr){4-5}\cmidrule(lr){6-7}\cmidrule(lr){8-9}
$K$
& \textsc{OC}$\uparrow${\scriptsize(\%)} & Time$\downarrow${\scriptsize(ms)}
& \textsc{OC}$\uparrow${\scriptsize(\%)} & Time$\downarrow${\scriptsize(ms)}
& \textsc{OC}$\uparrow${\scriptsize(\%)} & Time$\downarrow${\scriptsize(ms)}
& {\cellcolor{white!25}\textsc{OC}$\uparrow${\scriptsize(\%)}} & {\cellcolor{white!25}Time$\downarrow${\scriptsize(ms)}} \\
\midrule
4 & 98.9{\color{gray}\scriptsize$\pm$0.1} & 14.5{\color{gray}\scriptsize$\pm$3.9} & 68.4{\color{gray}\scriptsize$\pm$3.8} & 36.6{\color{gray}\scriptsize$\pm$7.3} & 46.9{\color{gray}\scriptsize$\pm$37.5} & 7.7{\color{gray}\scriptsize$\pm$2.6} & 77.6{\color{gray}\scriptsize$\pm$23.1} & 7.9{\color{gray}\scriptsize$\pm$2.6} \\
8 & 99.0{\color{gray}\scriptsize$\pm$0.1} & 17.7{\color{gray}\scriptsize$\pm$4.7} & 56.3{\color{gray}\scriptsize$\pm$11.0} & 39.8{\color{gray}\scriptsize$\pm$9.4} & 26.5{\color{gray}\scriptsize$\pm$32.4} & 12.9{\color{gray}\scriptsize$\pm$4.0} & 95.9{\color{gray}\scriptsize$\pm$10.8} & 13.1{\color{gray}\scriptsize$\pm$4.0} \\
16 & 99.2{\color{gray}\scriptsize$\pm$0.4} & 23.1{\color{gray}\scriptsize$\pm$5.1} & 90.5{\color{gray}\scriptsize$\pm$1.0} & 39.6{\color{gray}\scriptsize$\pm$8.4} & 14.3{\color{gray}\scriptsize$\pm$0.0} & 23.5{\color{gray}\scriptsize$\pm$6.3} & 100.0{\color{gray}\scriptsize$\pm$0.0} & 23.4{\color{gray}\scriptsize$\pm$6.3} \\
32 & 99.1{\color{gray}\scriptsize$\pm$0.4} & 32.4{\color{gray}\scriptsize$\pm$6.7} & 93.6{\color{gray}\scriptsize$\pm$0.8} & 43.5{\color{gray}\scriptsize$\pm$9.9} & 14.3{\color{gray}\scriptsize$\pm$0.0} & 43.4{\color{gray}\scriptsize$\pm$12.5} & 98.0{\color{gray}\scriptsize$\pm$5.4} & 43.4{\color{gray}\scriptsize$\pm$12.6} \\
\midrule
Overall & 99.0{\color{gray}\scriptsize$\pm$0.3} & 21.9{\color{gray}\scriptsize$\pm$8.5} & 77.2{\color{gray}\scriptsize$\pm$16.7} & 39.9{\color{gray}\scriptsize$\pm$8.7} & 25.5{\color{gray}\scriptsize$\pm$27.0} & 21.9{\color{gray}\scriptsize$\pm$15.6} & 92.9{\color{gray}\scriptsize$\pm$15.3} & 22.0{\color{gray}\scriptsize$\pm$15.5} \\
\bottomrule
\end{tabular}
\end{adjustbox}
\end{table}

\begin{table}[!h]
\centering
\caption{\textit{Ordered completion} and task-inference time by supplied \textit{view resolution} $V$ with dense task specifications. Both error terms describe variation across configuration means. $^{*}$\texttt{SR} and \texttt{SSR} with bisection goals.}
\vskip 0.1in
\label{tab:dense-view-resolution}
\renewcommand{\arraystretch}{1.15}
\setlength{\tabcolsep}{4pt}
\begingroup
\footnotesize
\renewcommand{\arraystretch}{1.15}
\setlength{\tabcolsep}{3pt}
\begin{tabular*}{\textwidth}{@{\extracolsep{\fill}\hspace{\tabcolsep}}ccccc@{\hspace{\tabcolsep}}}
\toprule
\multicolumn{3}{l}{\textbf{Experiment Config}} & \multicolumn{2}{l}{} \\
\midrule
$K^{\dagger}$ & Task input & Views & $N_{\mathrm{cfg}}$ & $N_{\mathrm{ep}}$ \\
\midrule
$\{4,8,16,32\}$ & Dense & \makecell[c]{$V=3$\\$1$ configuration} & $4\times1$ & $4\mathrm{k}$ \\
$\{4,8,16,32\}$ & Dense & \makecell[c]{$V=4$\\$2$ configurations$^{\dagger}$} & $4\times2$ & $8\mathrm{k}$ \\
$\{4,8,16,32\}$ & Dense & \makecell[c]{$V=5$\\$3$ configurations$^{\dagger}$} & $4\times3$ & $12\mathrm{k}$ \\
$\{4,8,16,32\}$ & Dense & \makecell[c]{$V=8$\\$1$ configuration} & $4\times1$ & $4\mathrm{k}$ \\
\bottomrule
\end{tabular*}
\endgroup
\par\smallskip
\begin{adjustbox}{width=\textwidth}
\begin{tabular}{c|cc|cc|cc|>{\columncolor{gray!25}}c>{\columncolor{gray!25}}c}
\toprule
\multicolumn{3}{l}{\textbf{Results}} & \multicolumn{6}{l}{} \\
\midrule
& \multicolumn{2}{c|}{\makecell[c]{\texttt{Skill Retrieval}\\(\texttt{SR})$^{*}$}}
& \multicolumn{2}{c|}{\makecell[c]{\texttt{Skill Sequence}\\\texttt{Retrieval} (\texttt{SSR})$^{*}$}}
& \multicolumn{2}{c|}{\makecell[c]{\texttt{Skill Chaining}\\(\texttt{SC})}}
& \multicolumn{2}{c}{\makecell[c]{\ours{}\\(ours)}} \\
\cmidrule(lr){2-3}\cmidrule(lr){4-5}\cmidrule(lr){6-7}\cmidrule(lr){8-9}
$V$
& \textsc{OC}$\uparrow${\scriptsize(\%)} & Time$\downarrow${\scriptsize(ms)}
& \textsc{OC}$\uparrow${\scriptsize(\%)} & Time$\downarrow${\scriptsize(ms)}
& \textsc{OC}$\uparrow${\scriptsize(\%)} & Time$\downarrow${\scriptsize(ms)}
& {\cellcolor{white!25}\textsc{OC}$\uparrow${\scriptsize(\%)}} & {\cellcolor{white!25}Time$\downarrow${\scriptsize(ms)}} \\
\midrule
3 & 99.0{\color{gray}\scriptsize$\pm$0.3} & 15.5{\color{gray}\scriptsize$\pm$6.1} & 75.3{\color{gray}\scriptsize$\pm$19.8} & 29.6{\color{gray}\scriptsize$\pm$1.4} & 14.3{\color{gray}\scriptsize$\pm$0.0} & 15.0{\color{gray}\scriptsize$\pm$10.6} & 75.0{\color{gray}\scriptsize$\pm$24.4} & 15.2{\color{gray}\scriptsize$\pm$10.4} \\
4 & 99.1{\color{gray}\scriptsize$\pm$0.2} & 19.1{\color{gray}\scriptsize$\pm$6.8} & 76.0{\color{gray}\scriptsize$\pm$18.5} & 35.1{\color{gray}\scriptsize$\pm$2.3} & 19.6{\color{gray}\scriptsize$\pm$21.5} & 18.5{\color{gray}\scriptsize$\pm$12.6} & 94.6{\color{gray}\scriptsize$\pm$15.2} & 18.4{\color{gray}\scriptsize$\pm$12.3} \\
5 & 99.0{\color{gray}\scriptsize$\pm$0.3} & 22.7{\color{gray}\scriptsize$\pm$7.2} & 76.4{\color{gray}\scriptsize$\pm$17.7} & 40.7{\color{gray}\scriptsize$\pm$2.8} & 22.6{\color{gray}\scriptsize$\pm$19.7} & 22.1{\color{gray}\scriptsize$\pm$14.6} & 95.2{\color{gray}\scriptsize$\pm$11.1} & 22.3{\color{gray}\scriptsize$\pm$14.7} \\
8 & 99.2{\color{gray}\scriptsize$\pm$0.2} & 31.8{\color{gray}\scriptsize$\pm$10.1} & 84.0{\color{gray}\scriptsize$\pm$10.1} & 57.3{\color{gray}\scriptsize$\pm$4.8} & 57.1{\color{gray}\scriptsize$\pm$49.5} & 34.9{\color{gray}\scriptsize$\pm$24.6} & 100.0{\color{gray}\scriptsize$\pm$0.0} & 35.0{\color{gray}\scriptsize$\pm$24.5} \\
\bottomrule
\end{tabular}
\end{adjustbox}
\end{table}

\subsubsection{Sub-trajectory Retrieval under Reduced Task Evidence}

To assess how sub-trajectory retrieval depends on task-observation density, we reduce the retained task evidence while keeping the stored procedure trajectories dense.
The adaptation matches task segments to stored sub-trajectories using subsequence dynamic time warping (S-DTW) and executes the skills associated with the retrieved evidence.
S-DTW adds linear interpolation of a sparse task query onto its original temporal grid before matching.
From one $383$-frame task demonstration, we uniformly retain observations while preserving the first and final observations and their timestamps.
Both methods receive the same interpolated task record as a dense task specification, from which \ours{} selects task observations by its task-query abstraction.
For procedure evidence, S-DTW uses dense, unsegmented trajectories, whereas \ours{} uses sparse initiation and termination observations organized into procedure records under $V=8$ views.
Table~\ref{tab:sdtw-retention} varies the number of retained observation steps.

\begin{table}[!h]
\centering
\small
\renewcommand{\arraystretch}{1.15}
\setlength{\tabcolsep}{3pt}
\caption{OC as fewer task observations are retained before temporal reconstruction. Task demo specs denote the number of retained observation steps.}
\vskip 0.1in
\label{tab:sdtw-retention}
\begingroup
\footnotesize
\renewcommand{\arraystretch}{1.15}
\setlength{\tabcolsep}{3pt}
\begin{tabular*}{\textwidth}{@{\extracolsep{\fill}\hspace{\tabcolsep}}ccccc@{\hspace{\tabcolsep}}}
\toprule
\multicolumn{3}{l}{\textbf{Experiment Config}} & \multicolumn{2}{l}{} \\
\midrule
$K^{\dagger}$ & $H$ & Views & $N_{\mathrm{cfg}}$ & $N_{\mathrm{ep}}$ \\
\midrule
$\{4,8,16,32\}$ & Dense & \makecell[c]{\ours{}: $V=8$\\S-DTW: --} & $4\times1$ & $4\mathrm{k}$ \\
\bottomrule
\end{tabular*}
\endgroup
\par\smallskip
\begin{tabular*}{\textwidth}{@{\extracolsep{\fill}\hspace{\tabcolsep}}l|rrrrrrrrr@{\hspace{\tabcolsep}}}
\toprule
\multicolumn{3}{l}{\textbf{Results}} & \multicolumn{7}{l}{} \\
\midrule
Task demo specs & 383 & 192 & 96 & 48 & 24 & 12 & 6 & 3 & 2 \\
\midrule
\ours{} & 100.0 & 100.0 & 100.0 & 100.0 & 100.0 & 100.0 & 100.0 & 100.0 & 100.0 \\
S-DTW & 100.0 & 85.7 & 85.7 & 96.4 & 92.9 & 57.1 & 28.6 & 0.0 & 0.0 \\
\bottomrule
\end{tabular*}
\end{table}

Both methods achieve $100.0\%$ OC with the complete task record.
S-DTW varies nonmonotonically at intermediate retention levels and reaches $0.0\%$ with three or two retained observation steps, whereas \ours{} remains at $100.0\%$ throughout this experiment.
This result concerns reconstruction from timestamped observations of this demonstration, with method-specific task processing applied afterward.

\subsubsection{Graph Search over Stored Trajectories}

To assess trajectory-graph composition under sparse task specifications, we vary the waypoint gap while retaining the same dense procedure trajectories.
We adapt the offline-state graph and shortest-path search of Test-Time Graph Search~\cite{opryshko2025test}, a mechanism also used for subgoal selection in Graph-Assisted Stitching~\cite{baek2025graph}.
Demonstration states form graph nodes, temporal edges preserve observed order, and nearest-neighbor edges connect states across trajectories.
Our adaptation uses squared Euclidean state distances and fixed temporal-step and stitching penalties, then applies Dijkstra's algorithm between supplied waypoints and executes the skills along the resulting path.
We set the cross-trajectory edge penalty to $2$, twice the temporal-step cost.
For the task specification, both methods receive the same sparse waypoints without interpolation at each $H\in\{1,\ldots,7\}$.
For procedure evidence, graph search uses dense, unsegmented trajectories without procedure views, while \ours{} uses sparse procedure-record endpoints under $V=8$ views.
The comparison in Table~\ref{tab:graph-search-gap} therefore differs in procedure evidence and composition under the same task inputs.

\begin{table}[!h]
\centering
\small
\renewcommand{\arraystretch}{1.15}
\setlength{\tabcolsep}{3pt}
\caption{OC across waypoint gaps for graph search.}
\vskip 0.1in
\label{tab:graph-search-gap}
\begingroup
\footnotesize
\renewcommand{\arraystretch}{1.15}
\setlength{\tabcolsep}{3pt}
\begin{tabular*}{\textwidth}{@{\extracolsep{\fill}\hspace{\tabcolsep}}ccccc@{\hspace{\tabcolsep}}}
\toprule
\multicolumn{3}{l}{\textbf{Experiment Config}} & \multicolumn{2}{l}{} \\
\midrule
$K^{\dagger}$ & $H$ & Views & $N_{\mathrm{cfg}}$ & $N_{\mathrm{ep}}$ \\
\midrule
$\{4,8,16,32\}$ & $\{1,\ldots,7\}$ & \makecell[c]{\ours{}: $V=8$\\Others: --} & $4\times7$ & $28\mathrm{k}$ \\
\bottomrule
\end{tabular*}
\endgroup
\par\smallskip
\begin{tabular*}{\textwidth}{@{\extracolsep{\fill}\hspace{\tabcolsep}}l|rrrrrrr@{\hspace{\tabcolsep}}}
\toprule
\multicolumn{3}{l}{\textbf{Results}} & \multicolumn{5}{l}{} \\
\midrule
Method / $H$ & 1 & 2 & 3 & 4 & 5 & 6 & 7 \\
\midrule
\ours{} & 100.0 & 100.0 & 100.0 & 100.0 & 100.0 & 100.0 & 100.0 \\
Graph search & 46.4 & 14.3 & 7.1 & 0.0 & 0.0 & 0.0 & 0.0 \\
\bottomrule
\end{tabular*}
\end{table}

Graph-search OC falls from $46.4\%$ at $H=1$ to $7.1\%$ at $H=3$ and $0.0\%$ for $H\geq4$.
\ours{} remains at $100.0\%$ at every gap.

To isolate the role of procedure support in graph search, we fix $H=1$ and vary skill-context diversity in Table~\ref{tab:graph-search-diversity}.
The graph construction and edge penalty of $2$ remain fixed.

\begin{table}[!h]
\centering
\small
\renewcommand{\arraystretch}{1.15}
\setlength{\tabcolsep}{3pt}
\caption{Graph-search OC across skill-context diversity at $H=1$.}
\vskip 0.1in
\label{tab:graph-search-diversity}
\begingroup
\footnotesize
\renewcommand{\arraystretch}{1.15}
\setlength{\tabcolsep}{3pt}
\begin{tabular*}{\textwidth}{@{\extracolsep{\fill}\hspace{\tabcolsep}}ccccc@{\hspace{\tabcolsep}}}
\toprule
\multicolumn{3}{l}{\textbf{Experiment Config}} & \multicolumn{2}{l}{} \\
\midrule
$K$ & $H$ & Views & $N_{\mathrm{cfg}}$ & $N_{\mathrm{ep}}$ \\
\midrule
$\{4,8,16,32\}$ & $1$ & -- & $4\times1$ & $4\mathrm{k}$ \\
\bottomrule
\end{tabular*}
\endgroup
\par\smallskip
\begin{tabular*}{\textwidth}{@{\extracolsep{\fill}\hspace{\tabcolsep}}lrrrr@{\hspace{\tabcolsep}}}
\toprule
\multicolumn{3}{l}{\textbf{Results}} & \multicolumn{2}{l}{} \\
\midrule
Skill-context diversity $K$ & $4$ & $8$ & $16$ & $32$ \\
\midrule
Graph search & 100.0 & 42.9 & 42.9 & 0.0 \\
\bottomrule
\end{tabular*}
\end{table}

Graph search achieves $100.0\%$ OC at the smallest diversity setting and lower completion at the larger settings.
Thus, adding procedure demonstrations does not improve completion in this construction.
The aggregate results alone do not identify which graph connections cause the errors.

}
}

%% file: neurips/checklist.tex
\section*{NeurIPS Paper Checklist}

\begin{enumerate}

\item {\bf Claims}
    \item[] Question: Do the main claims made in the abstract and introduction accurately reflect the paper's contributions and scope?
    \item[] Answer: \answerYes{} %
    \item[] Justification: The abstract and introduction explicitly state the motivation and contributions.

\item {\bf Limitations}
    \item[] Question: Does the paper discuss the limitations of the work performed by the authors?
    \item[] Answer: \answerYes{} %
    \item[] Justification: We outline future work that aims to address current limitations.

\item {\bf Theory assumptions and proofs}
    \item[] Question: For each theoretical result, does the paper provide the full set of assumptions and a complete (and correct) proof?
    \item[] Answer: \answerNA{} %
    \item[] Justification: The paper does not include theoretical results. 

    \item {\bf Experimental result reproducibility}
    \item[] Question: Does the paper fully disclose all the information needed to reproduce the main experimental results of the paper to the extent that it affects the main claims and/or conclusions of the paper (regardless of whether the code and data are provided or not)?
    \item[] Answer: \answerYes{} %
    \item[] Justification: We provide a clear specification of the framework components, the experimental settings used in our evaluation, and the corresponding evaluation protocol.

\item {\bf Open access to data and code}
    \item[] Question: Does the paper provide open access to the data and code, with sufficient instructions to faithfully reproduce the main experimental results, as described in supplemental material?
    \item[] Answer: \answerYes{} %
    \item[] Justification: The code, data-generation scripts, and evaluation scripts will be released on GitHub with instructions to reproduce the main experimental results.

\item {\bf Experimental setting/details}
    \item[] Question: Does the paper specify all the training and test details (e.g., data splits, hyperparameters, how they were chosen, type of optimizer) necessary to understand the results?
    \item[] Answer: \answerYes{} %
    \item[] Justification: We specify the experimental settings and implementation details in Section 6 of the main paper and provide additional details in the appendix. 

\item {\bf Experiment statistical significance}
    \item[] Question: Does the paper report error bars suitably and correctly defined or other appropriate information about the statistical significance of the experiments?
    \item[] Answer: \answerYes{} %
    \item[] Justification: We report our results with variance.

\item {\bf Experiments compute resources}
    \item[] Question: For each experiment, does the paper provide sufficient information on the computer resources (type of compute workers, memory, time of execution) needed to reproduce the experiments?
    \item[] Answer: \answerYes{} %
    \item[] Justification: We provide these details in the experiments section and the appendix.
    
\item {\bf Code of ethics}
    \item[] Question: Does the research conducted in the paper conform, in every respect, with the NeurIPS Code of Ethics \url{https://neurips.cc/public/EthicsGuidelines}?
    \item[] Answer: \answerYes{} %
    \item[] Justification: We adhere to the NeurIPS Code of Ethics in every respect.

\item {\bf Broader impacts}
    \item[] Question: Does the paper discuss both potential positive societal impacts and negative societal impacts of the work performed?
    \item[] Answer: \answerYes{} %
    \item[] Justification: We discuss the potential impacts in the conclusion section.
    
\item {\bf Safeguards}
    \item[] Question: Does the paper describe safeguards that have been put in place for responsible release of data or models that have a high risk for misuse (e.g., pre-trained language models, image generators, or scraped datasets)?
    \item[] Answer: \answerNA{} %
    \item[] Justification: The paper does not release high-risk models or datasets such as pre-trained language models, image generators, or Internet-scraped datasets. Our experiments are conducted in a controlled benchmark. Therefore, no additional misuse-specific release safeguards are applicable.

\item {\bf Licenses for existing assets}
    \item[] Question: Are the creators or original owners of assets (e.g., code, data, models), used in the paper, properly credited and are the license and terms of use explicitly mentioned and properly respected?
    \item[] Answer: \answerYes{} %
    \item[] Justification: We credit the creators and original owners of the existing assets used in our experiments.

\item {\bf New assets}
    \item[] Question: Are new assets introduced in the paper well documented and is the documentation provided alongside the assets?
    \item[] Answer: \answerYes{} %
    \item[] Justification: The released code includes documentation for running the experiments and reproducing the reported results.

\item {\bf Crowdsourcing and research with human subjects}
    \item[] Question: For crowdsourcing experiments and research with human subjects, does the paper include the full text of instructions given to participants and screenshots, if applicable, as well as details about compensation (if any)? 
    \item[] Answer: \answerNA{} %
    \item[] Justification: The paper does not involve crowdsourcing nor research with human subjects.

\item {\bf Institutional review board (IRB) approvals or equivalent for research with human subjects}
    \item[] Question: Does the paper describe potential risks incurred by study participants, whether such risks were disclosed to the subjects, and whether Institutional Review Board (IRB) approvals (or an equivalent approval/review based on the requirements of your country or institution) were obtained?
    \item[] Answer: \answerNA{} %
    \item[] Justification: The paper does not involve crowdsourcing nor research with human subjects.

\item {\bf Declaration of LLM usage}
    \item[] Question: Does the paper describe the usage of LLMs if it is an important, original, or non-standard component of the core methods in this research? Note that if the LLM is used only for writing, editing, or formatting purposes and does \emph{not} impact the core methodology, scientific rigor, or originality of the research, declaration is not required.
    \item[] Answer: \answerNA{} %
    \item[] Justification: The core method development in this research does not involve LLMs as any important, original, or non-standard components.

\end{enumerate}